\documentclass[10pt,journal, compsoc]{IEEEtran}
\usepackage{times}
\usepackage{epsfig}
\usepackage{graphicx}
\usepackage{amsmath}
\usepackage{amssymb}
\usepackage{subfig}
\usepackage{array}
\usepackage{multirow}
\usepackage{colortbl}
\usepackage[table]{xcolor}
\newcolumntype{P}[1]{>{\centering\arraybackslash}p{#1}}
\newcolumntype{M}[1]{>{\centering\arraybackslash}m{#1}}
\definecolor{TableRed}{HTML}{800000}
\newcommand{\TextRed}[1]{\textcolor{TableRed}{#1}}
\ifCLASSOPTIONcompsoc
  \usepackage[nocompress]{cite}
\else
  \usepackage{cite}
\fi
\ifCLASSINFOpdf
\else
\fi
\newcommand{\etal}{\textit{et al}}

\begin{document}
\title{Differential Viewpoints for Ground Terrain Material Recognition}

\author{Jia~Xue, Hang~Zhang, Ko~Nishino, and ~Kristin~J.~Dana% <-this % stops a space
\IEEEcompsocitemizethanks{
\IEEEcompsocthanksitem Jia Xue is  with the Department of   Electrical and Computer Engineering, Rutgers University--New Brunswick, New Brunswick, NJ 08901, USA. \protect
E-mail: jia.xue@rutgers.edu

\IEEEcompsocthanksitem Hang Zhang is with the Amazon AI, Amazon Web Services Inc, East Palto Alto, CA 94025, USA
\protect
E-mail: hzaws@amazon.com

\IEEEcompsocthanksitem Ko Nishino is with the Department of Intelligence Science and Technology, Graduate School of Informatics, Kyoto University
Yoshida Honmachi, Sakyo-ku Kyoto, Kyoto 606-8501. \protect
E-mail: kon@i.kyoto-u.ac.jp

\IEEEcompsocthanksitem Kristin J. Dana is with the   Department   of   Electrical   and   Computer Engineering, Rutgers University--New Brunswick, New Brunswick, NJ 08901,USA. \protect
E-mail: kristin.dana@rutgers.edu}% <-this % stops a space

% \thanks{Manuscript received April 19, 2005; revised August 26, 2015.}
}

\IEEEtitleabstractindextext{%
\begin{abstract}
%Material recognition for real-world outdoor surfaces has become increasingly important for computer vision to support its operation ``in the wild.'' %for instance, for their applications to automated driving, robotics and human-computer interaction. 
Computational surface modeling that underlies material recognition has transitioned from reflectance modeling using in-lab controlled radiometric measurements to image-based representations based on internet-mined single-view images captured in the scene. We take a middle-ground approach for material recognition that takes advantage of both rich radiometric cues and flexible image capture.  %accomplishing material recognition by in-scene multiple view image sequence comprising more information than a single photograph, but less information than a lab-based reflectance measurement.
A key concept is differential angular imaging, where small angular variations in image capture  enables angular-gradient features for an enhanced appearance representation that improves recognition.  
We build a large-scale material database, Ground Terrain in Outdoor Scenes (GTOS) database, to support ground terrain recognition for applications such as autonomous driving and robot navigation. The database consists of over 30,000 images covering 40 classes of outdoor ground terrain under varying weather and lighting conditions.
We develop a novel approach for material recognition called texture-encoded angular network (TEAN) that combines deep encoding pooling of RGB information and differential angular images for angular-gradient features to fully leverage this large dataset. With this novel network architecture, we extract characteristics of materials encoded in the angular and spatial gradients of their appearance. Our results show that TEAN achieves recognition performance that surpasses single view performance and standard (non-differential/large-angle sampling) multiview performance. 
%These results demonstrate the effectiveness of differential angular imaging as a means for flexible, in-place material recognition. 
\end{abstract}

}

\maketitle

% is not selected <OR> if conference mode is selected - because compsoc
% conference papers position the abstract like regular (non-compsoc)
\IEEEdisplaynontitleabstractindextext

% \fi
\IEEEpeerreviewmaketitle

\ifCLASSOPTIONcompsoc
\IEEEraisesectionheading{\section{Introduction}\label{sec:introduction}}
\else
\section{Introduction}
\label{sec:introduction}
\fi

\begin{figure}[h]
\centering
\includegraphics[width= .9\linewidth]{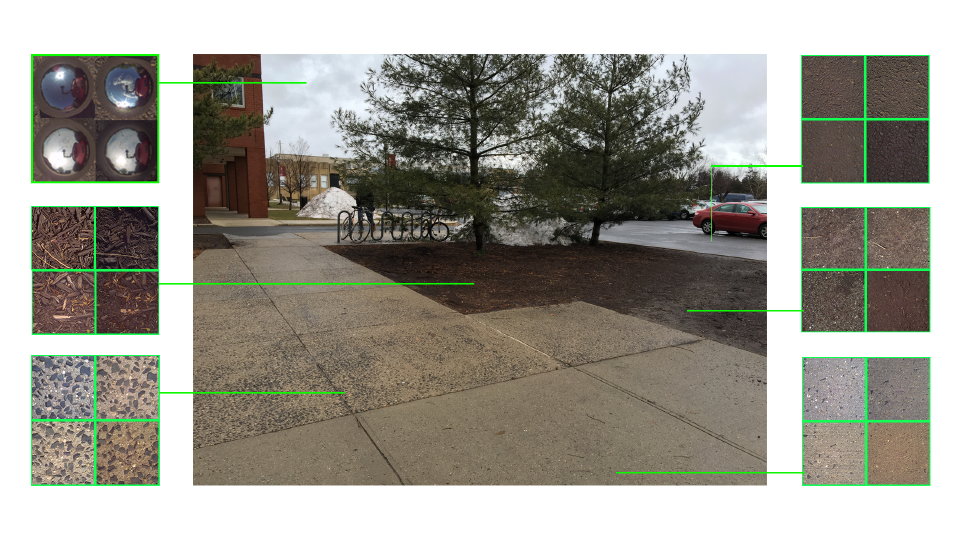}
\includegraphics[width= .9\linewidth]{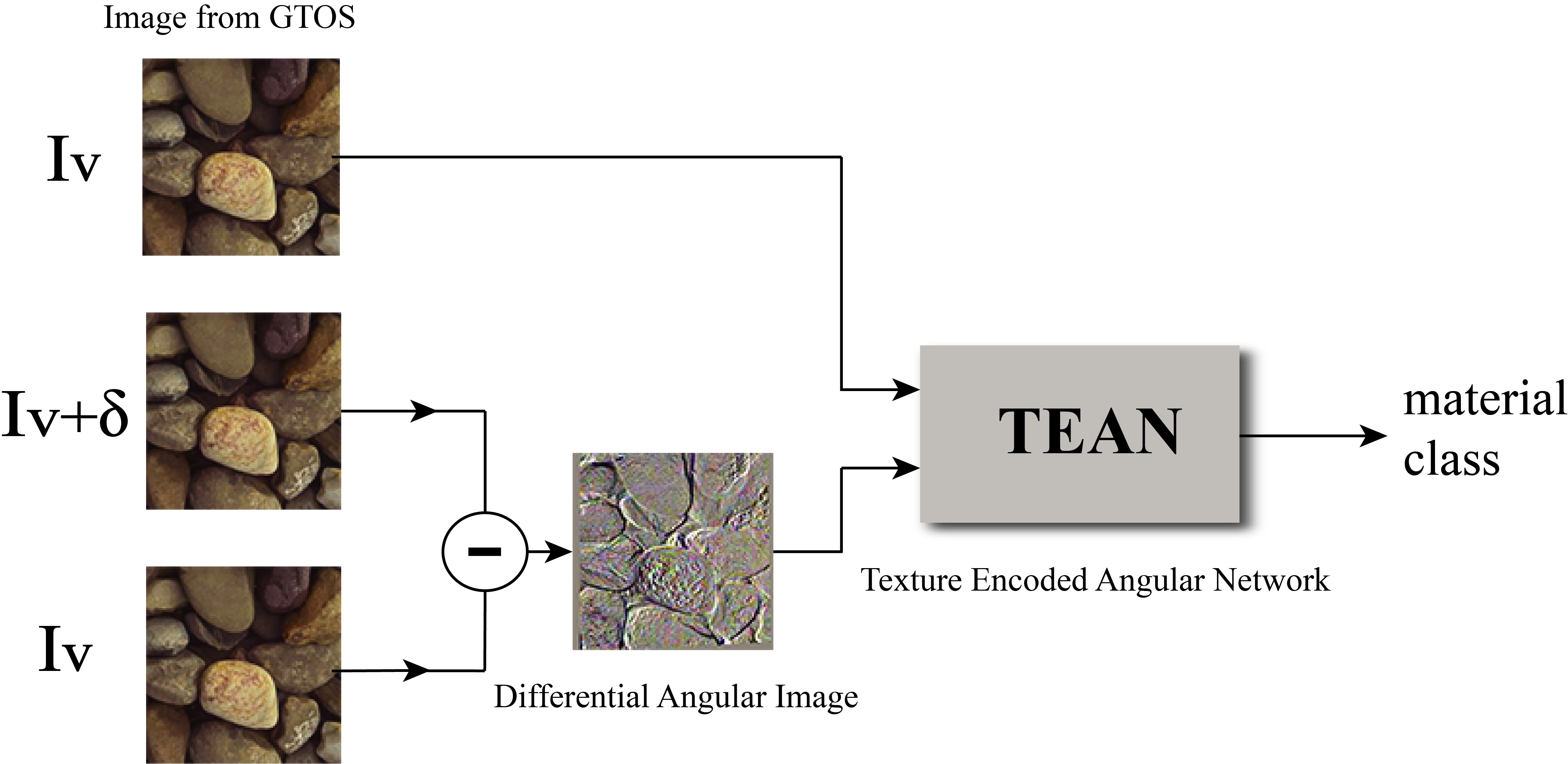}
\caption{(Top) Example from GTOS dataset comprising   outdoor measurements with multiple viewpoints, illumination conditions and angular differential imaging.  The example shows scene-surfaces imaged  at different illumination/weather conditions.  (Bottom)  Texture Encoded Angular Network (TEAN) for ground terrain material recognition. }
\label{fig:figure1}
\end{figure}

\IEEEPARstart{R}{eal} world scenes consist of surfaces made of numerous materials, such as wood, marble, dirt, metal, ceramic and fabric, which contribute to the rich visual variation we find in images. 
Material recognition has become an active area of research in recent years, with the goal of providing detailed material information for applications such as autonomous agents and human-machine systems.
Real world surfaces and material characteristics are both apparent (the visual appearance) and latent (physical material properties of the surface such as friction, micro-geometry and roughness).
%Modeling the apparent and latent characteristics of different materials is essential to robustly recognize them in images. 
Material properties affect both the spatial variation of surface appearance and the angular variation of reflectance with respect to both view and illumination.
% An important question to  address is how multiple surface views (from different angles) of the surface enhance material recognition. 

\begin{figure*}[h]
\centering
\subfloat
{
\includegraphics[width=.13\linewidth]{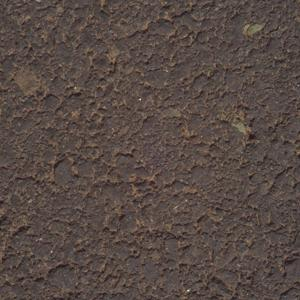}
}
\subfloat
{
\includegraphics[width=.13\linewidth]{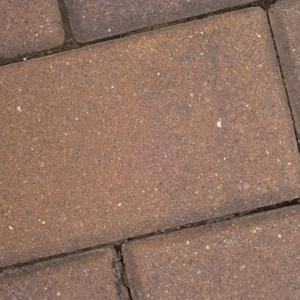}
}
\subfloat
{
\includegraphics[width=.13\linewidth]{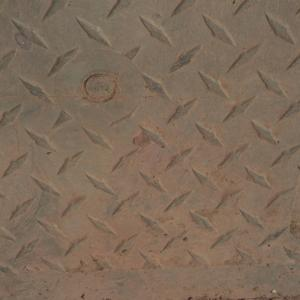}
}
\subfloat
{
\includegraphics[width=.13\linewidth]{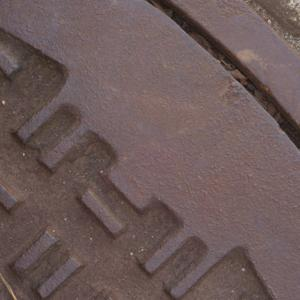}
}
\subfloat
{
\includegraphics[width=.13\linewidth]{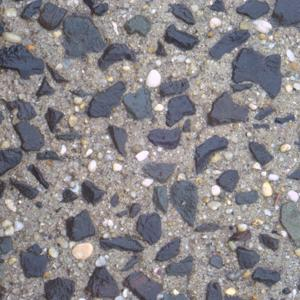}
}
\subfloat
{
\includegraphics[width=.13\linewidth]{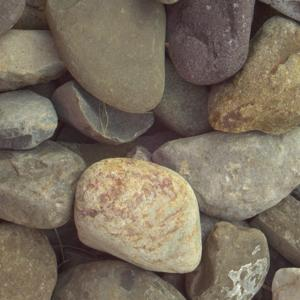}
}
\subfloat
{
\includegraphics[width=.13\linewidth]{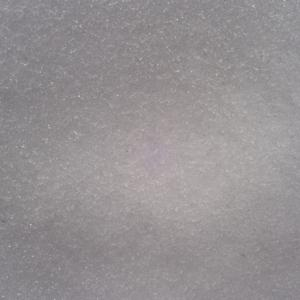}
}
\setcounter{subfigure}{0}
\subfloat[Asphalt]
{
\includegraphics[width=.13\linewidth]{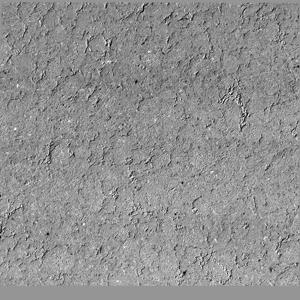}
}
\subfloat[Brick]
{
\includegraphics[width=.13\linewidth]{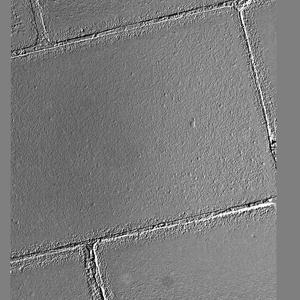}
}
\subfloat[Plastic cover]
{
\includegraphics[width=.13\linewidth]{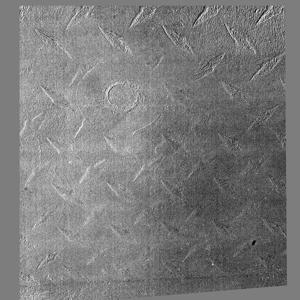}
}
\subfloat[Metal cover]
{
\includegraphics[width=.13\linewidth]{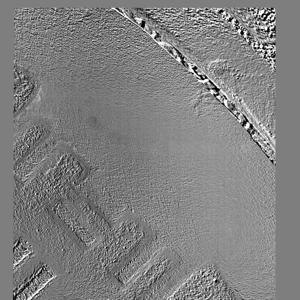}
}
\subfloat[Stone-cement]
{
\includegraphics[width=.13\linewidth]{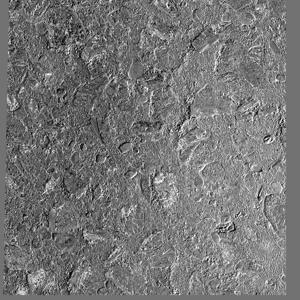}
}
\subfloat[Pebble]
{
\includegraphics[width=.13\linewidth]{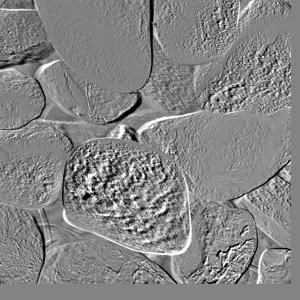}
}
\subfloat[Snow]
{
\includegraphics[width=.13\linewidth]{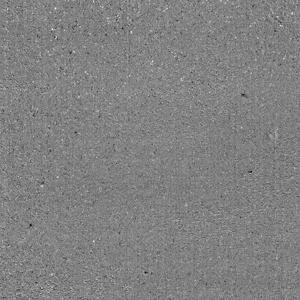}
}
\caption{Differential Angular Imaging. (Top) Examples of material surface images $I_v$. (Bottom) Corresponding  differential images $I_\delta = I_v - I_{v+\delta}$ in our GTOS dataset. These sparse images encode angular gradients of reflection and 3D relief texture.}
\label{fig:compare_img_patch}
\end{figure*}

Early studies of material appearance modeling largely concentrated on comprehensive lab-based measurements using dome systems, robots, or gonioreflectometers collecting measurements that are dense in angular space (such as BRDF, BTF) \cite{dana2016capturing}. These reflectance-based studies have the advantage of capturing intrinsic invariant properties of the surface, which enables fine-grained material recognition \cite{Liu_PAMI14,Salamati_CIC09,Wang_CVPR09,zhang2015reflectance,li2018materials}. The inflexibility of lab-based image capture, however, prevents widespread use in real world scenes, especially in the important class of outdoor scenes. 
A fundamentally different approach to reflectance modeling is image-based appearance modeling where surfaces are captured with a single-view image in-scene or ``in-the-wild.'' 
Recent studies of image-based material recognition use single-view internet-mined images to train classifiers \cite{bell15minc,cimpoi2015deep,zhang2016deep,bu2019deep,andrearczyk2016using,kong2017low} and can be applied to arbitrary images casually taken without the need of multiview reflectance information. 
%Surface regions imaged in the context of the scene are labeled with the corresponding material categories. 
In these methods, however, recognition is typically based more on context including object and scene cues, than intrinsic material appearance properties except for a few purely local methods \cite{schwartz2013visual,schwartz2015automatically}.

Between the two approaches of reflectance-based and image-based material recognition, i.e.\ between comprehensive in-lab imaging and internet-mined images, we take an advantageous middle-ground. 
Specifically, we capture in-scene real-world surfaces but use multiple viewpoint angles for measurements that provide a partial reflectance sampling. This leads to a very basic question: how do multiple viewing angles help in material recognition?  
More interestingly, we consider a novel question:  Do small changes in viewing angles, {\it differential changes}, result in significant increases in recognition performance?
Prior work has shown the power of angular filtering to complement spatial filtering in material recognition. These methods, however, rely on a lightfield camera to achieve multiple differential viewpoint variations \cite{wang20164d} or a mirror-based camera to capture a slice of the BRDF \cite{zhang2015reflectance}   which limits application in the wild due to a rigid imaging system setup and  inadequacy for image capture at distance.
We instead propose to capture surfaces with differential changes in viewing angles with an ordinary camera and compute discrete approximations of {\it angular gradients}.
We present an approach called {\it angular differential imaging} that augments image 
capture for a particular viewing angle $v$ a differential viewpoint $v+\delta$. Contrast this method with lab-based reflectance measurements that often quantize the angular space measuring with domes or positioning devices with large angular spacing such as $22.5^\circ$. These coarse-quantized measurements have limited use in approximating angular gradients. 
Angular differential imaging can be implemented with a small-baseline stereo camera or a moving camera (e.g. handheld).
We demonstrate that differential angular imaging provides key information about material reflectance properties while maintaining the flexibility of convenient in-scene appearance capture. 

To capture material appearance in a manner that preserves the convenience of image-based methods but important angular information of reflectance-based methods, we assemble a comprehensive, first-of-its-kind, {\it outdoor, in-place} material database that includes multiple viewpoints and multiple illumination directions (partial BRDF sampling), multiple weather conditions, a large set of surface material classes surpassing existing comparable datasets, multiple physical instances per surface class (to capture intra-class variability) and differential viewpoints to support the framework of differential angular imaging.  
Specifically, the resulting database spans 40 surface classes that we find commonly in daily life, 4-14 examples or instances per class and for each surface measurement we collect an image set at 18  viewing angles using a mobile robot and at multiple illumination conditions (4) corresponding to times of day and weather conditions. Each image is collected with 3 exposures for high dynamic range imaging. The total number of surface images is 34,243. These surfaces are not measured in the lab, but rather in their typical state within a scene. The global scene image of the surface is also captured and indicates the scene context for the surface. %  (See Figure~\ref{fig:figure1}).  
%category              40
%material region     197
%material surfaces  606
%material images    34243
We concentrate on outdoor scenes because of the limited availability of reflectance databases for outdoor surfaces. We also concentrate on materials from ground terrain in outdoor scenes (GTOS) for applicability in numerous application such as automated driving, robot navigation and scene semantics. The 40 surface classes include ground terrain such as grass, gravel, asphalt, concrete, black ice, snow, moss, mud and sand (see Figure~\ref{fig:compare_img_patch}).

%build a overview of the recognition methods, to leverage the introduced differential angular image with the color image, we employ a two branch network to combine both. THen we find that for the color branch, we can utilize both spatial and texture information. Finally, we combine both information for the final network.

For recognition, we build a recognition algorithm that leverages the strength of deep learning and differential angular imaging. We develop a two-branch network that combines deep encoding pooling for spatial (texture) information and a second branch for angular information as illustrated in Figure~\ref{fig:figure1}. We call this new architecture Texture Encoded Angular Network (TEAN). It combines two prior concepts DEP network\cite{xue2018deep} and DAIN \cite{xue2016differential} to account for spatial texture and angular information in a robust deep learning architecture. The original concepts have been improved with new architectures to incorporate better base networks with improved efficiency and accuracy. 

%We build a recognition algorithm that leverages the strength of deep learning and differential angular imaging. The resulting method takes two image streams as input, the original image and a differential image as illustrated in Figure~\ref{fig:figure1}.We optimize the two-stream configuration for material recognition performance and call the resulting network DAEN--deep angular encoding network. 

\section{Related Work}

\begin{table*}[t]
\centering

\begin{tabular}{|l|P{1cm}|P{1cm}|P{.9cm}|P{1.8cm}|P{1.4cm}|P{1.9cm}|P{1.6cm}|l|}
\hline 

Datasets & samples & classes & views & illumination & in scene &scene image& camera parameters& year\\

\hline 
CUReT\cite{dana1999reflectance}&61&61&\multicolumn{2}{P{2.7cm}|}{205}&N&N&N&1999\\

\hline 
KTH-TIPS\cite{hayman2004significance}&11&11&27&3&N&N&N&2004\\

\hline 
UBO2014\cite{weinmann2014material}&84&7&151&151&N&N&N&2014\\

\hline 
Reflectance disk\cite{zhang2015reflectance}&190&19&3&3&N&N&Y&2015\\

\hline 
4D Light-field\cite{wang20164d}&1200&12&1&1&Y&N&N&2016\\
\hline 
NISAR\cite{Choe16cvpr}&100&100&9&12&N&N&N&2016\\
\hline 

\textbf{GTOS(ours)}&\textbf{606}&\textbf{40}&\textbf{19}&\textbf{4}&\textbf{Y}&\textbf{Y}&\textbf{Y}&\textbf{2017} \\
         
\hline 
\end{tabular}
\caption{Comparison between GTOS dataset and some publicly available BRDF material datasets. Note that the 4D Light-field dataset\cite{wang20164d} is captured by the Lytro Illum light field camera.}
\label{table:dataset_comparison}
\end{table*}

{\bf Material recognition:} Material recognition is a fundamental problem in computer vision. The classification of 3D material images and bidirectional texture functions, traditionally relies on handcrafted filter banks followed by grouping the outputs into texton histograms \cite{Cula01,Varma2005} or bag-of-words \cite{Csurka04, Lazebnik06, liu2018survey, liu2019bow}. The success of deep learning methods in object recognition has also translated to the problem of material recognition, the classification and segmentation of material categories in arbitrary images. Bell \etal. achieve per-pixel material category labeling by retraining the state-of-the-art object recognition network \cite{simonyan2014two} on a large dataset of material appearance \cite{bell15minc}. This method relies on large image patches that include object and scene context to recognize materials. In contrast, Schwartz and Nishino \cite{schwartz2013visual,schwartz2015automatically} learn material appearance models from small image patches extracted inside object boundaries to decouple contextual information from material appearance. To achieve accurate local material recognition, they introduced intermediate material appearance representations based on their intrinsic properties (e.g., ``smooth'' and ``metallic''). Zhang \etal  \cite{zhang2016deep} introduce Deep Texture Encoding Network (Deep-TEN) that ports the dictionary learning and feature pooling approaches into the CNN pipeline for an end-to-end material/texture recognition network that learns an encoding for an orderless texture representation. 
These prior methods show the utility of spatial information within an image for material recognition.  
%Zhi \etal \cite{zhi2019multispectral} use multi-spectral imaging together with visible light and near infrared for powder classification.

While there has been recent emphasis of characterizing materials with apparent appearance in images,  radiometric properties of materials such as the bidirectional reflectance distribution function (BRDF) \cite{united1977geometrical} and the bidirectional texture function (BTF) \cite{dana1999reflectance} provide appearance as a function of viewing and illumination angle and therefore encode angular information. Materials have unique characteristics in the subtle variations of their reflectance functions  (e.g., different types of metal \cite{liu2014discriminative} and paint \cite{wang2009material}). However, reflectance measurements require elaborate image capture systems, such as a gonioreflectometer \cite{united1977geometrical, ward1992measuring}, robotic arm \cite{levoy1996light, dana1999reflectance}, or a dome with cameras and light sources \cite{debevec2000acquiring, liu2014discriminative, wang2009material}. (Numerous methods for capturing reflectance have been detailed in surveys \cite{dana2016capturing, Filip09}.)
 Recently, Zhang \etal\ introduced the use of a one-shot reflectance field capture for material recognition \cite{zhang2015reflectance}. They adapt the parabolic mirror-based camera developed by Dana and Wang \cite{dana2004device} to capture the BRDF for a given light source direction in a single shot, called a reflectance disk. These results demonstrate that gradients of angular appearance  encode rich cues for their recognition. Similarly, Wang \etal \cite{wang20164d} uses a light field camera and combines angular and spatial filtering for material recognition. The approach we present in this paper  develops a novel material recognition framework that combines  both spatial and angular filtering. 
 Specifically, we combine state-of-the-art texture representations with reflectance cues from differential angular images that can easily be captured by a two-camera system or small motions of a single ordinary camera. The resulting method is instantiated in a two-branch network comprised of one branch for reflectance with an emphasis on angular gradients and another branch for texture with both orderless and ordered spatial cues. 
 
 %however, we use standard cameras instead of a multilens array such as Lytro. Also, we capture large viewing angle range (with samples coarsely quantized in angle space) and using differential changes around each angle. The differential images 

{\bf Datasets:} Datasets to measure reflectance of real world surfaces have a long history of lab-based measurements including: CUReT database\cite{dana1999reflectance}, KTH-TIPS database by Hayman \etal. \cite{hayman2004significance}, MERL Reflectance Database \cite{Matusik03}, UBO2014 BTF Database \cite{weinmann2014material}, UTIA BRDF Database \cite{Filip14}, Drexel Texture Database \cite{Oxholm_ECCV12-2} and IC-CERTH Fabric Database \cite{Kampouris16}.
In many of these datasets, dense reflectance angles are captured with special image capture equipment. Some of these datasets have limited instances/samples per surface category (different physical samples representing the same class for intraclass variability) or have few surface categories, and all are obtained from indoor measurements where the sample is removed from the scene. 
More recent datasets capture materials and texture in-scene, (a.k.a. in-situ, or in-the-wild). A motivation of moving to in-scene capture is to build algorithms and methods that are more relevant to real-world applications. 
These recent databases are from internet-mined databases and contain a single view of the scene under a single illumination direction. Examples include the the Flickr Materials Database by Sharan \etal. \cite{sharan2009material} and the Material in Context Database by Bell \etal. \cite{bell15minc}. However, because photographs in both datasets are collected from internet, the reflectance properties from multiple views 
%and camera parameters
are lost. 
Recent work uses deep networks to estimate multiview reflectance models  for novel-view material rendering \cite{deschaintre2018single, li2018learning, li2018materials, gao2019deep, deschaintre2019flexible, riviere2017polarization}. Training with multiview renderings may be a future direction for cases where real-world images are not available.

\section{GTOS Dataset}

\begin{figure}
\centering
\includegraphics[width= \linewidth]{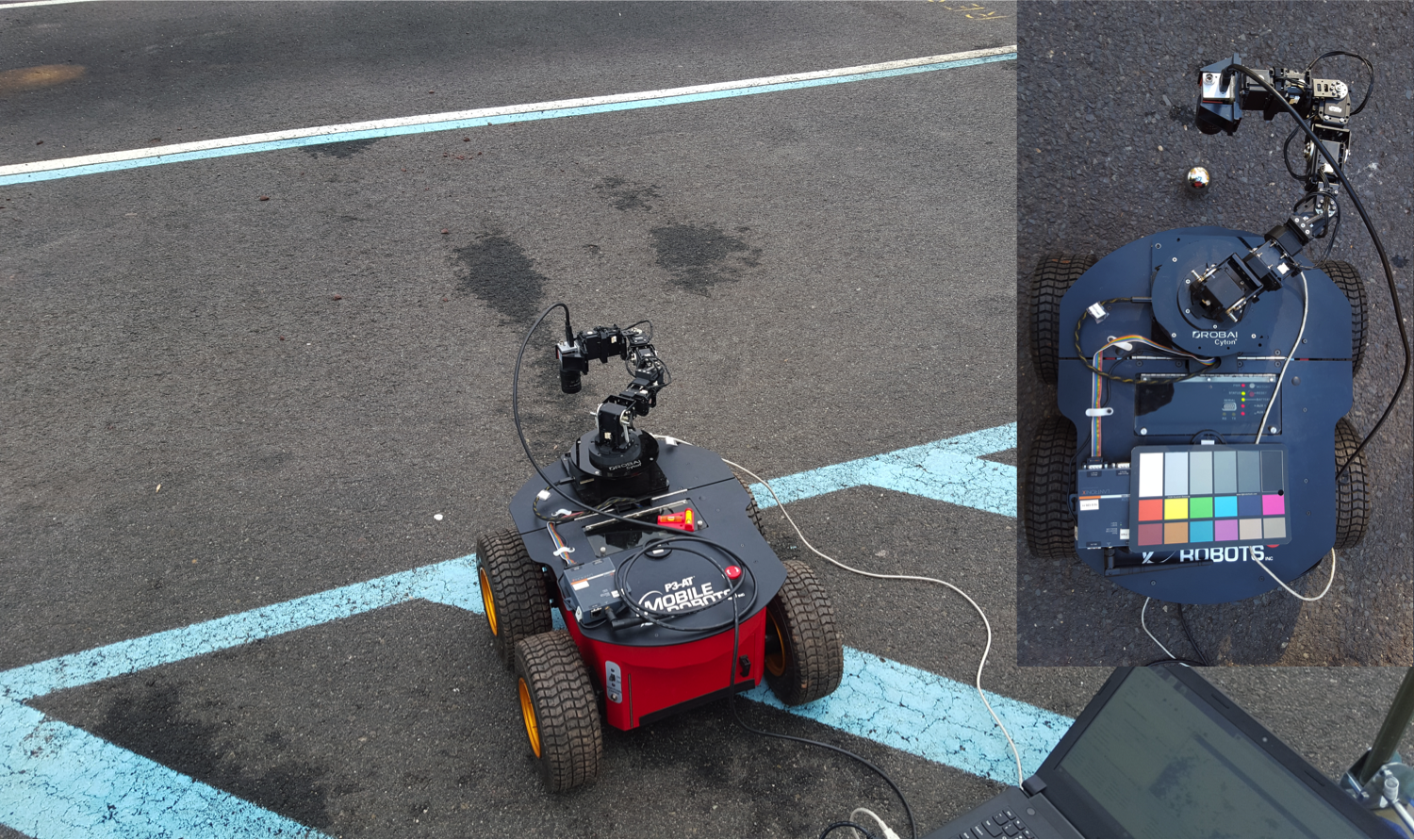}
\caption{The measurement equipment for the GTOS database: Mobile Robots P3-AT robot, Cyton gamma 300 robot arm, Basler aca2040-90uc camera with Edmund Optics 25mm/F1.8 lens, DGK 18\% white balance and color reference card, and the 440C Stainless Steel Tight-Tolerance Sphere (McMaster-Carr).}
\label{fig:equipment}
\end{figure}

\begin{figure*}[t]
\centering
\subfloat[material classes]
{
  \includegraphics[width=.3\linewidth]{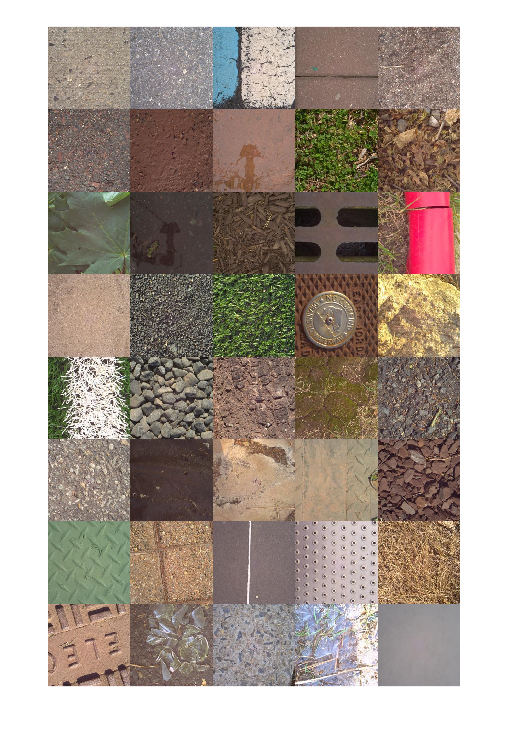}
  
}
\subfloat[one sample at multiple viewing directions]
{
  \includegraphics[width=.65\linewidth]
  {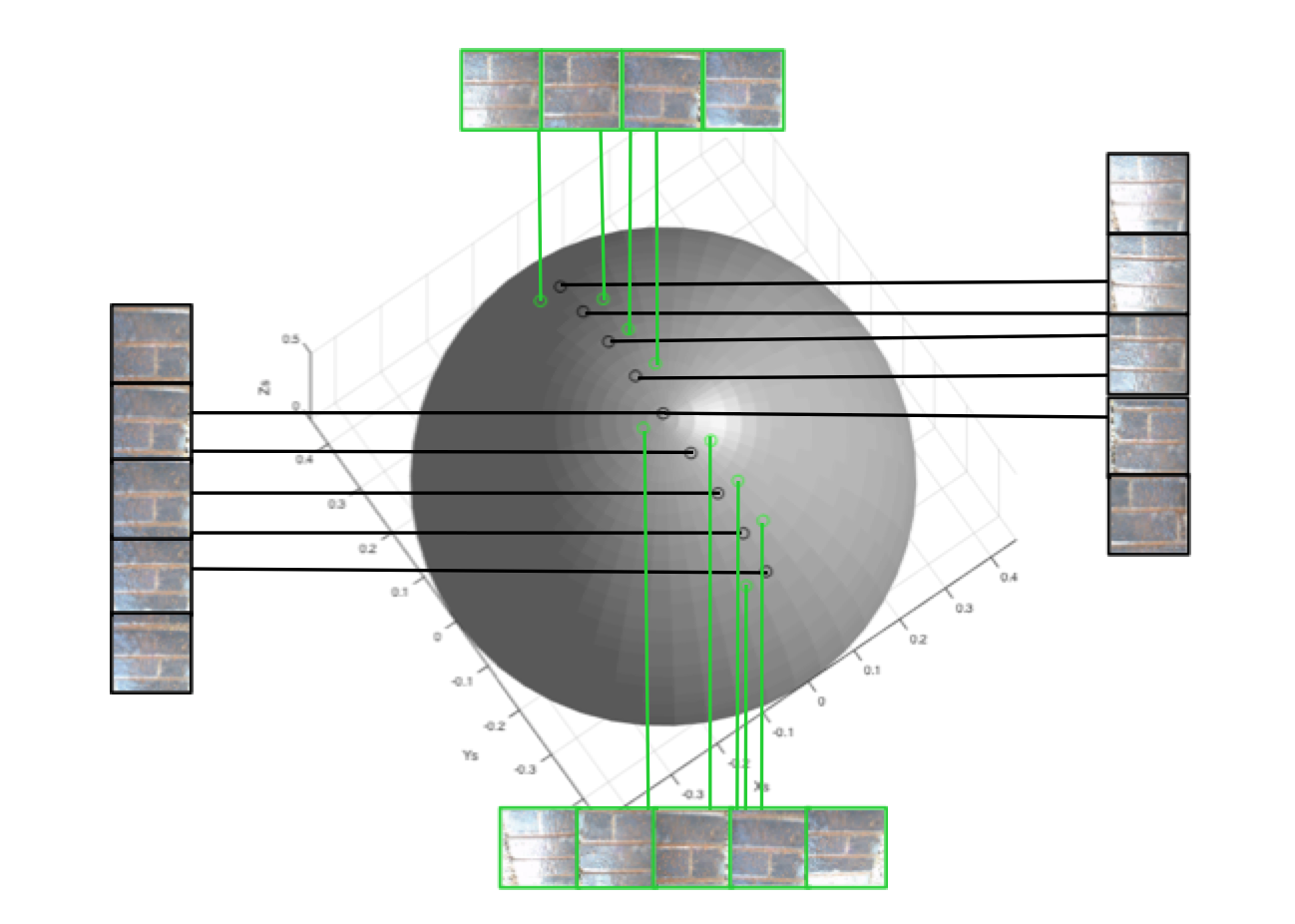}
}
\caption{(a) The 40 material categories in the GTOS dataset introduced in this paper. (b) The material surface observation points. Nine viewpoint angles (black spots) separated along an arc spanning $80^\circ$ are measured. For each viewpoint, a differential view (green spots) is captured $\pm5^\circ$ in azimuth from the original orientation (the sign is chosen based on robotic arm kinematics. )}
\label{fig:database}
\end{figure*}

In this section, we introduce the GTOS dataset and the measurement device. GTOS dataset is a first-of-its-kind in-scene material reflectance database, to investigate the use of spatial and angular reflectance information of outdoor ground terrain for material recognition.

\subsection{Measurement Device}
Our measurement device  (depicted in Figure~\ref{fig:equipment})  is composed of a Mobile Robots P3-AT robot, Cyton gamma 300 robot arm, Basler aca2040-90uc camera with Edmund Optics 25mm/F1.8 lens, DGK 18\% white balance and color reference card, and Hardened 440C Stainless Steel Tight-Tolerance Sphere (McMaster-Carr). The constraint that the Cyton arm can only hold 300g for full-range movement presents a practical obstacle in our choice for camera and lens, and we employed the Basler USB camera with Edmund Industrial optics 86572 fixed focus lens (the total weight of the camera and the lens is 203g). The aca2040-90uc camera can capture 2040 $\times$ 2046 pixels resolution photographs with 12 bits per pixel. As shown in Figure~\ref{fig:database}, a Hardened 440C Stainless Steel Tight-Tolerance Sphere is employed to reflect the sky and indicate the weather conditions. Camera parameter adjustment is challenging due to image capture in different lighting. We set the camera parameters to be adjusted automatically by simultaneously observing the DGK 18\% white balance and color reference card. Camera parameters are adjusted only one time for each sample, ensuring that images at multiple viewing angles are captured under the same parameters. 
Sample appearance depends on sky/weather conditions and the time of day.  We image the same region with four different weather conditions (cloudy dry, cloudy wet, sunny morning and sunny afternoon).
As shown in Figure~\ref{fig:database}, we choose 9 points to form an approximate $80^\circ$ arc as our viewing points.
 % Each view point has a 3 to 5 degree variance alpha point.
For each observation image, an additional image obtained by varying the viewing angle by a small angle ($3-5^\circ$) provides the pair needed to compute the angular gradient. 
To collect BRDF information, the observation points are fixed for the entire database. The distance between observing points and sample is 4045 mm. The imaging region is 1510 mm $\times$ 1510 mm. 

\subsection{Dataset Overview}
We collect the GTOS database, a first-of-its-kind in-scene material reflectance database, to investigate the use of spatial and angular reflectance information of outdoor ground terrain for material recognition. 
We capture reflectance systematically by imaging a set of viewing angles comprising a partial BRDF with a mobile exploration robot.
The measurement device is depicted in Figure~\ref{fig:equipment}. Due to the joint limitation of the Cyton gamma 300 robot arm, we select $[-40^\circ,40^\circ]$ as our measuring range.
Differential angular images are obtained by measuring each of $N_v=9$ base angles $v=(\theta_v,\phi_v)$, $\theta_v \in [-40^\circ,-30^\circ,\dots,40^\circ]$, and a differential angle variation of $\delta = (0,5^\circ)$ resulting in 18 viewing directions per sample as shown in Figure~\ref{fig:database} (b).
Example surface classes are depicted in Figure~\ref{fig:database} (a).
The class names are (in order of top-left to bottom-right): cement, asphalt, painted asphalt , brick, soil, muddy stone, mud, mud-puddle, grass, dry leaves, leaves, asphalt-puddle, mulch, metal grating, plastic, sand, stone, artificial turf, aluminum,  limestone, painted turf, pebbles, roots, moss, loose asphalt-stone, asphalt-stone, cloth, paper, plastic cover, shale, painted cover, stone-brick, sandpaper, steel, dry grass, rusty cover, glass, stone-cement, icy mud, and snow. The $N_c=40$ surface classes mostly have between 4 and 14 instances (samples of intra-class variability) and each instance is imaged not only under $N_v$ viewing directions but also under multiple natural light illumination conditions. As illustrated in Figure~\ref{fig:figure1}, sample appearance depends on the weather condition and the time of day.  
To capture this variation, we image the same region with $N_i=4$ different weather conditions (cloudy dry, cloudy wet, sunny morning, and sunny afternoon).
We capture the samples with 3 different exposure times to enable high dynamic range imaging. Additionally, we image a mirrored sphere
to capture the environment lighting of the natural sky. 
In addition to surface images, we capture a scene image to show the global context. Although, the database measurements were obtained with robotic positioning for precise angular measurements, our recognition results are based on subsets of these measurements so that an articulated arm would not be required for an in-field system.
% add footnote
The total number of surface images in the database is 34,243. 
As shown in Table~\ref{table:dataset_comparison}, this is the most extensive outdoor in-scene multiview material database to date.

\subsection{Differential Angular Imaging}
Our GTOS dataset introduces a  measurement method called differential angular imaging where a surface is imaged from a particular viewing angle $v$ and then from an additional viewpoint $v+\delta$. The motivation for this differential change in viewpoint is improved computation of the angular gradient of intensity $\partial{I_v}/\partial{v}$. Intensity gradients are the basic building block of image features and it is well known that discrete approximations to derivatives have limitations. In particular, spatial gradients of intensities for an image $I$ are approximated by $I(x+\Delta)-I(x)$ and this approximation is most reasonable at low spatial frequencies and when $\Delta$ is small. 
%. In particular, consider a general function $f(x)$ with   the continuous gradient $f'(x)$ approximated by a discrete difference $f(x+1) - f(x)$, the equivalent operation is convolution by a filter of the form $d(x) = h[x+\Delta] - h[x]$ where $h[x]$ denotes the discrete impulse function. From basic signal processing, this convolution operation is equivalent to the multiplication in the frequency domain by the discrete fourier transform of d(x) given by $D(\omega)$   The continuous gradient is an operation that is equivalent to multiplying by a filter response that increases linearly with frequency $\omega$. However  $D(\omega)$ only approximates the continuous gradient at low frequencies. 
One  implication is that the discrete approximation to the derivative is only valid at lower frequencies, as expected.
The second implication is that increasing $\Delta$ decreases the range of frequencies over which the discrete approximation is valid. Therefore, small values of $\Delta$ provide better gradients. 
For angular gradients of reflectance, the discrete approximation to the derivative is a subtraction with respect to the viewing angle. Angular gradients are approximated by $I(v+\delta)-I(v)$ and this approximation requires a small $\delta$. Consequently, differential angular imaging provides more accurate angular gradients.  

The differential angular images as shown in Figures~\ref{fig:figure1} and \ref{fig:compare_img_patch} have several characteristics. First, the differential angular image reveals the gradients in BRDF/BTF at the particular viewpoint. Second, relief texture is also observable in the differential angular image due to non-planar surface structure. Finally, the differential angular images are sparse. This sparsity has the potential to provide a computational advantage, though we have not specifically  utilized this advantage in our network design.

\section{Deep Learning Architectures}

\begin{figure*}[h]
\centering
\subfloat[Final layer (prediction) combination method]
{
  \includegraphics[width=.35\linewidth]{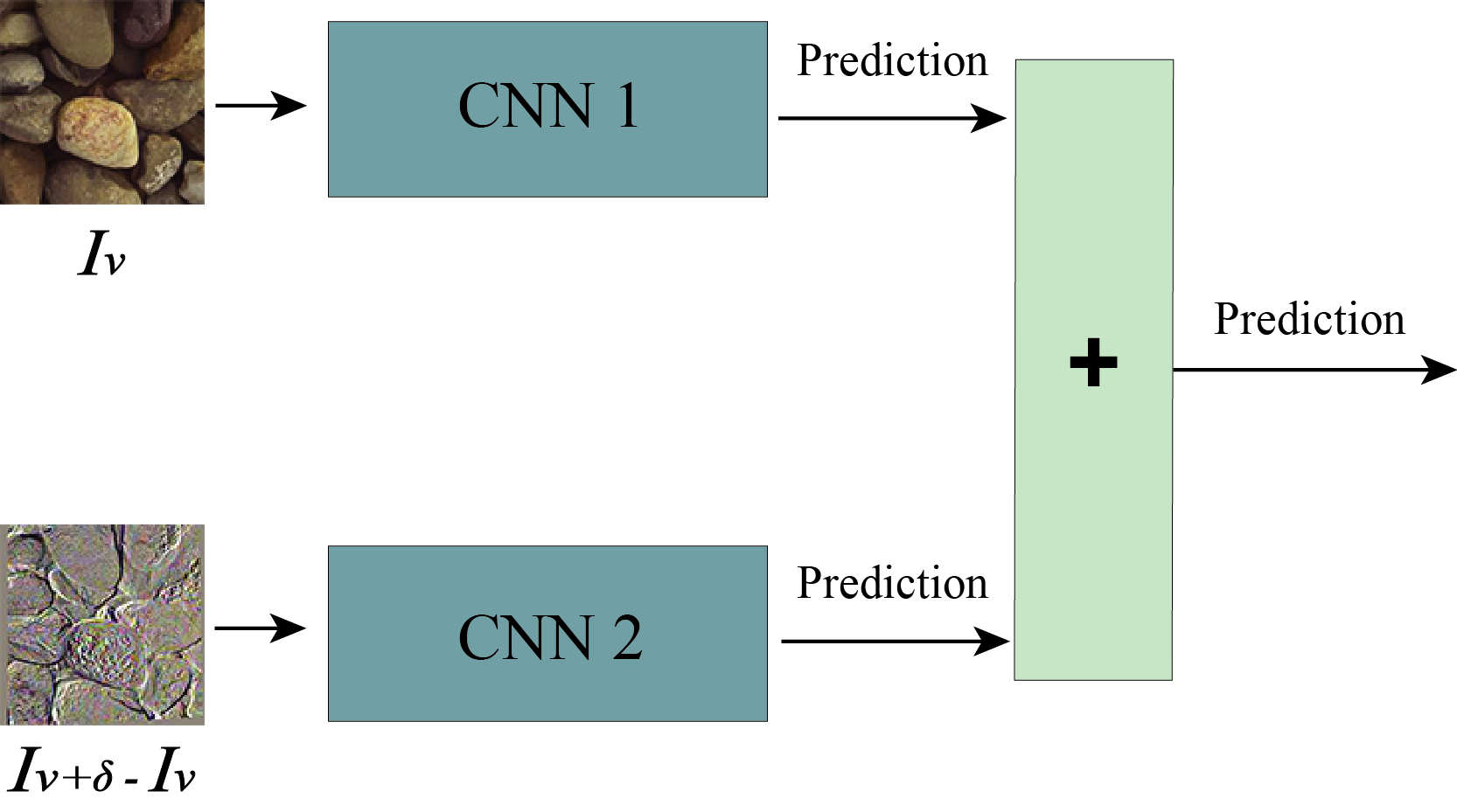}
  
}
\subfloat[Intermediate layer (feature maps) combination method]
{
  \includegraphics[width=.46\linewidth]
  {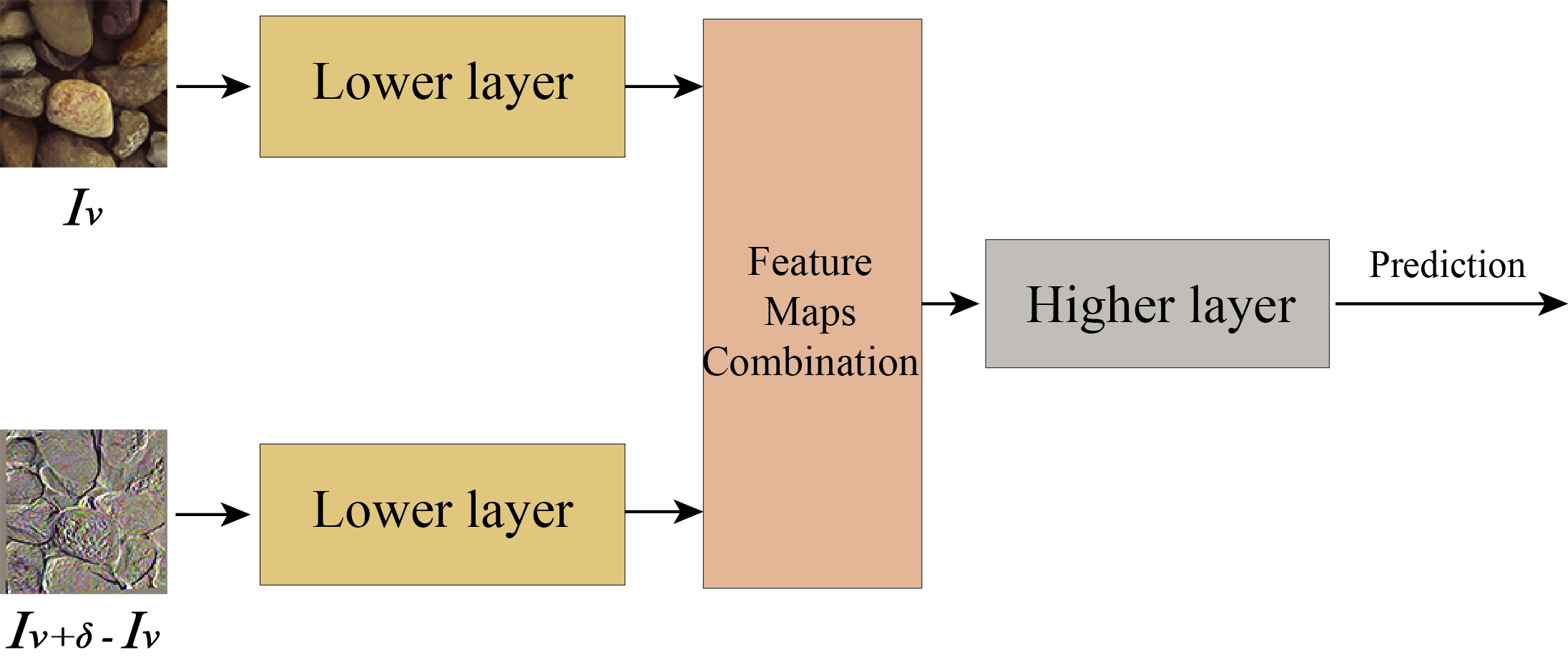}
}

\subfloat[Both intermediate layer and final layer combination]
{
  \includegraphics[width=.6\linewidth]
  {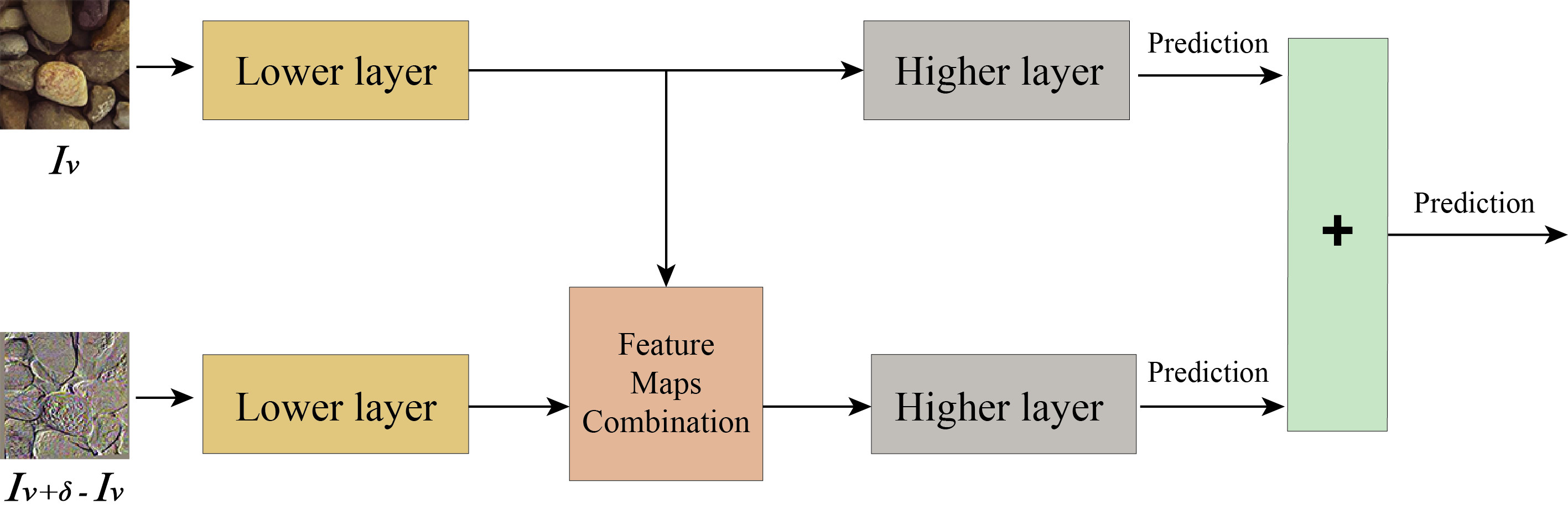}
}
\caption{Methods to combine two image streams, the original image $I_v$ and the differential angular image $I_\delta= I_{v+\delta}-I_v$. The architecture in (c)  provides better performance than (a) and (b) and we call it the differential angular imaging network (DAIN). }
\label{fig:fuse_cmp}
\end{figure*}

To leverage differential angular imaging for material recognition, we build a two stream convolution network that takes two image streams as input, the original image and a differential image. We start our experiments with widely studied ImageNet \cite{deng2009imagenet} pre-trained networks \cite{he2016deep, chatfield2014return, sandler2018mobilenetv2} as the CNN streams, and we call the network Differential Angular Imaging Network (DAIN). Networks designed for object recognition take spatial order as critical for classification. However, texture recognition uses an orderless component to provide invariance to spatial layout \cite{lin2013network,cimpoi2015deep}. Through study the GTOS dataset, we find that for ``images in the wild", homogeneous surfaces rarely fill the entire field-of-view, and many materials exhibit regular structure. %For texture recognition, since surfaces are not completely orderless, {\it local spatial order}  is an important cue for recognition. Just as semantic segmentation balances local details and global scene context for pixelwise recognition \cite{Zheng_2015_ICCV,bell2015material}, 
We design a network to balance both orderless  and ordered spatial information for the GTOS images, and we call this network Deep Encoding Pooling Network (DEP). Finally, by replacing the CNN architecture in DAIN  with DEP to combine angular reflectance cues with  orderless and ordered spatial infromation, we introduce  Texture Encoded Angular Network (TEAN).

\begin{figure*}[t]
\centering
\includegraphics[width= .95\textwidth]{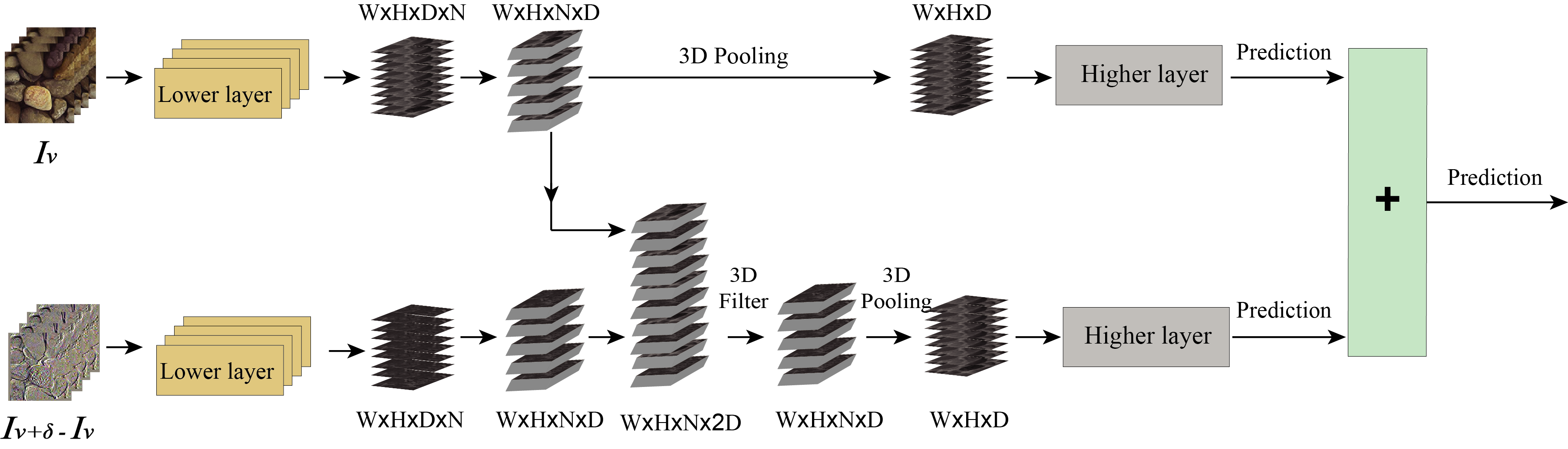}
\caption{Multiview DAIN. The 3D filter + pooling method to combine two streams (original and differential image) from multiple viewing angles
$v \in [v1,v2,...,vN]$.   $W$, $H$, and $D$ are the width, height, and depth of corresponding feature maps, $N$ is the number of view points.}
\label{fig:3D_filter}
\end{figure*}

\subsection{Differential Angular Imaging Network (DAIN)}
We develop a two-stream convolutional neural network to fully leverage differential angular imaging for material recognition. 
The differential angular image $I_\delta$ sparsely encodes reflectance angular gradients as well as surface relief texture. The spatial variation of image intensity remains an important recognition cue and so our method integrates these two streams of information. 
A CNN is used on both streams of the network and then combined for the final prediction result. The combination method and the layer at which the combination takes place leads to variations of the architecture.

We employ the ImageNet \cite{deng2009imagenet} pre-trained VGG-M model \cite{chatfield2014return} as the initial prediction unit (labeled CNN in Figure~\ref{fig:fuse_cmp}). The first input branch is the image  $I_v$ at a specific viewing direction $v$. The second input branch is the differential image $I_\delta$. 
The first method of combination  shown in Figure~\ref{fig:fuse_cmp} (a) is a simple averaging of the output prediction vectors obtained by the two branches. The second method  combines the two branches at the intermediate layers of the CNN, i.e. the feature maps output at layer $M$ are combined and passed forward to the higher layers of the CNN, as shown Figure~\ref{fig:fuse_cmp} (b).
We empirically find that combining feature maps generated by Conv5 layer after ReLU performs best.
A third method (see Figure~\ref{fig:fuse_cmp} (c)) is a hybrid of the  two architectures that preserves the original CNN path for the original image $I_v$ by combining the layer $M$ feature maps for both streams {\it and} by combining the prediction outputs for both streams as shown in Figure~\ref{fig:fuse_cmp} (c). This approach is the best performing architecture of the three methods and we call it the differential angular imaging network (DAIN). 

For combining feature maps at layer $M$,
consider features maps ${x}_{a}$ and $x_b$ from the two branches 
that have width $W$, height $H$, and feature channel depth $D$. 
The output feature map $y$ will be the same dimensions  ${W \times H \times D}$.
We can combine feature maps by: (1) {\it Sum:} pointwise sum of ${x}_{a}$ and $x_b$, and (2) {\it Max:} pointwise maximum of ${x}_{a}$ and $x_b$. As experimentally proved in section~\ref{DAIN_result}, the sum combination outperforms max combination. Without explicit declaration, we use sum combination as our default combination method.
The CNN module of our DAIN network can be replaced by other state-of-the-art deep learning methods to further improve results. To demonstrate this, in section~\ref{sec:experiments}, we change the VGG-M model with ImageNet pre-trained MobileNet V2 \cite{sandler2018mobilenetv2} and provide several experiments for evaluation.
%In Section~\ref{sec:experiments} we evaluate the performance of these methods of combining lower layer feature maps. 

\vspace{.1in}
\noindent\textbf{Multiple Views}
Our GTOS database has multiple viewing directions on an arc (a partial BRDF sampling) as well as differential images for each viewing direction. 
We evaluate  our recognition network in two modes:
(1) {\bf Single view DAIN}, with inputs from $I_v$ and $I_\delta$, with $v$ representing a single viewing angle; (2) {\bf Multiview DAIN}, with inputs $I_v$ and $I_\delta$, with $v \in [v1,v2,...,vN]$. For our GTOS database,  $v1,v2,...,vN$ are viewing angles separated by $10^\circ$ representing a $N \times 10^\circ$ range of viewing angles. We empirically determine that $N=4$ viewpoints are sufficient for recognition. 
For a baseline comparison we also consider non-differential versions: {\bf Single View} with only $I_v$ for a single viewing direction and {\bf Multiview} with inputs $I_v$,   $v \in [v1,v2,...,vN]$. 

To incorporate multiview information in DAIN we use three methods:
(1) voting (use the predictions from each view to vote), 
(2) pooling (pointwise maximum of the combined feature maps across viewpoints),
(3) 3D filter + pooling (follow \cite{tran2015learning} to use a $3\times 3 \times 3$ learned filter bank to convolve the multiview feature maps). See Figure~\ref{fig:3D_filter}. 
After 3D filtering, pooling is used (pointwise maximum across viewpoints). 
Due to learning the filter weights, the computational expense of this third method is significantly higher.

\subsection{Deep Encoding Pooling Network (DEP)}
\begin{figure*}[h]
\centering
\includegraphics[width= \linewidth]{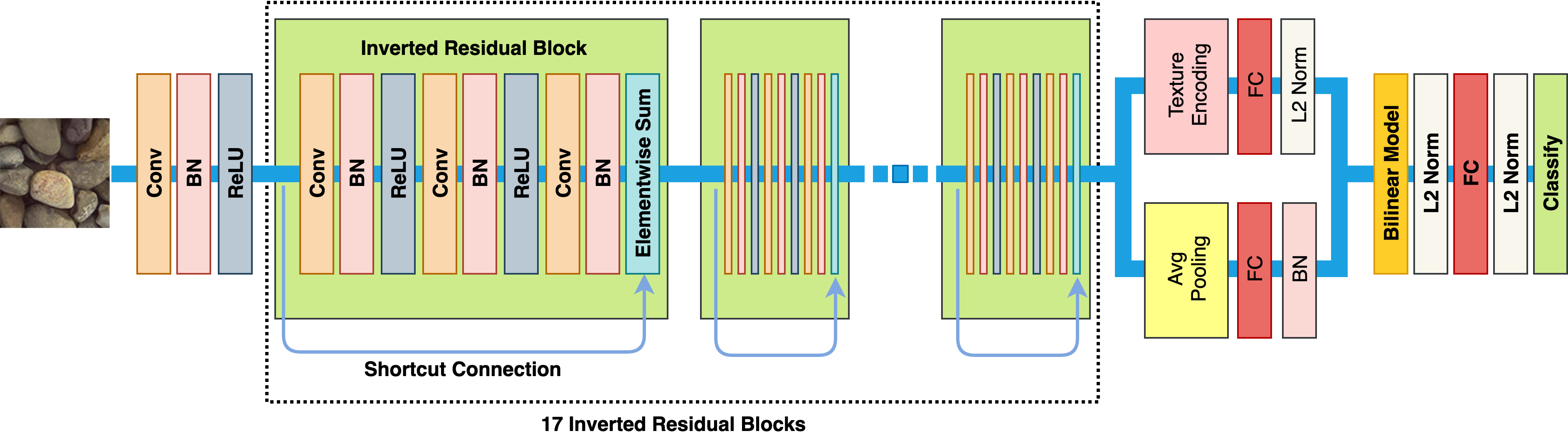}
\caption{A Deep Encoding Pooling Network (DEP) for material recognition. Outputs from convolutional layers are fed into  the encoding layer and global average pooling layer jointly and their outputs are processed with bilinear model.}
\label{fig:algorithm}
% \vspace{-0.2in}
\end{figure*}

We introduce a Deep Encoding Pooling Network (DEP) that leverages an orderless texture representation and high level spatial information for material recognition. The network is shown in Figure~\ref{fig:algorithm}, with the material image as network input, outputs from convolutional layers are fed into two feature representation layers jointly; the texture encoding layer \cite{zhang2016deep} and the global average pooling layer. The texture encoding layer captures texture appearance details and the global average pooling layer accumulates spatial information. Features from the encoding layer and the global average pooling layer are processed with bilinear models \cite{tenenbaum1997separating}.

\vspace{.1in}
\noindent\textbf{Encoding Layer} The texture encoding layer~\cite{zhang2016deep} integrates the entire dictionary learning and visual encoding pipeline into a single CNN layer, which provides an orderless representation for texture modeling. 
The encoding layer acts as a global feature pooling on top of convolutional layers. 
Here we briefly describe prior work for completeness. %is an orderless feature pooling layer that works as the dictionary learning and feature pooling in classic material recognition approaches.  %is integrated on top of convolution layers and represents  outputs from convolution layer with learnable codewords. 
Let $X = \left \{ {x}_{1},...{x}_{m} \right \}$ be M visual descriptors, $C = \left \{ {c}_{1},...{c}_{n} \right \}$ is the code book with N learned codewords. The residual vector ${r}_{ij}$ is calculated by ${r}_{ij} = {x}_{i} - {c}_{j}$, where $i = 1 ...m$ and $j = 1 ...n$.
The residual encoding for codeword ${c}_{j}$ can be represented as 
\begin{equation}
{e}_{j} = \sum_{i=1}^{M} {w}_{ij}{r}_{ij} ,
\end{equation}
where ${w}_{ij}$ is the assigning weight for residual vector ${r}_{ij}$ and is given by
\begin{equation}
{w}_{ij} = \frac{\exp(-{s}_{j}{\lVert{r}_{ij}\rVert}^{2})}{\sum_{k=1}^{m} \exp({-{s}_{k}{\lVert{r}_{ik}\rVert}^{2}})} ,
\end{equation}
$ {s}_{1},...{s}_{m}$ are learnable smoothing factors. With the texture encoding layer, the visual descriptors X are pooled into a set of N residual encoding vectors $E = \left \{ {e}_{1},...{e}_{n} \right \}$. 
% "we can find" is not very formal
Similar to classic encoders, the encoding layer can capture more texture details by increasing the number of learnable codewords.

\vspace{.1in}
\noindent\textbf{Bilinear Models}  
Bilinear models are two-factor models such that their outputs are linear in one factor if the other factor is constant~\cite{freeman1997learning}.
The factors in bilinear models balance the contributions of the two components. 
Let ${a}^{t}$ and ${b}^{s}$ represent the material texture information and spatial information with vectors of parameters and with dimensionality $I$ and $J$. The bilinear function $Y^{ts}$ is given by 
\begin{equation}
Y^{ts} = \sum_{i=1}^{I}\sum_{j=1}^{J} {w}_{ij}{a}_{i}^{t}{b}_{j}^{s} ,
\end{equation}
where ${w}_{ij}$ is a learnable weight to balance the interaction between material texture and spatial information.  The outer product representation captures a pairwise correlation between the material texture encodings and spatial observation structures.

\vspace{.1in}
\noindent\textbf{Deep Encoding Pooling Network}
With aforementioned encoding layer and bilinear models, we introduce our Deep Encoding Pooling Network (DEP). Our Deep  Encoding Pooling Network is shown in Figure~\ref{fig:algorithm}. As in prior transfer learning algorithms \cite{lin2015bilinear,zhang2016deep}, we employ  convolutional layers with non-linear layers from ImageNet \cite{deng2009imagenet} pre-trained CNNs as feature extractors.  Outputs from convolutional layers  are fed into the texture encoding layer and the global average pooling layer jointly. Outputs from the texture encoding layer preserve  texture details, while outputs from the global average pooling layer preserve high level spatial information. The dimension of outputs from the texture encoding layer is determined by the codewords N and the feature maps channel C (N$\times$C). The  dimension of outputs from the global average pooling layer is determined by the  feature maps channel C. For computational efficiency and to robustly combine feature maps with bilinear models, we reduce feature maps dimension with fully connected layers for both branches. Feature maps from the texture encoding layer and the global average pooling layer are processed with a bilinear model and followed by a fully connected layer and a classification layer with non-linearities for classification. Table ~\ref{table:structure} is an instantiation of DEP based on MobileNet V2 \cite{sandler2018mobilenetv2}. We set 8 codewords for the texture encoding layer. The size of input images are $224\times224$. Outputs from CNNs are fed into the texture encoding layer and the global average pooling layer jointly. The dimension of outputs from the texture encoding layer is $8\times1280=10240$ and the dimension of outputs from global average pooling layer is 1280.
We reduce the dimension of feature maps from the texture encoding layer and the global average pooling layer to 64 via fully connected layers. The dimension of outputs from bilinear model is $64\times64=4096$.
Following prior works \cite{zhang2016deep,perronnin2010improving}, resulting vectors from the texture encoding layer and bilinear model are normalized with L2 normalization.

The texture encoding layer and bilinear models are both differentiable.
The overall architecture is a directed acyclic graph and all the parameters can be trained by back propagation.
Therefore, the Deep Encoding Pooling Network is trained end-to-end using stochastic gradient descent with back-propagation.

{\setlength{\extrarowheight}{3pt}%
\begin{table}
\centering
\resizebox{\linewidth}{!}{%
\begin{tabular}{|l|c|c|}
\hline
layer name & output size & encoding-pooling \\ \hline
conv2d      & 112$\times$112$\times$32  & 3$\times$3, stride 2    \\ 
\hline
bottleneck1\_x       & 112$\times$112$\times$16   &  $\begin{bmatrix} 3 \times 3, 32\\ 1 \times 1, 16 \end{bmatrix} \times 1$    \\ [1em]
\hline
bottleneck2\_x     & 56$\times$56$\times$24   &  $\begin{bmatrix} 1 \times 1, 96\\3 \times 3, 96\\ 1 \times 1, 24 \end{bmatrix} \times 2$    \\ [1em]
\hline 
bottleneck3\_x     & 28$\times$28$\times$32   &  $\begin{bmatrix} 1 \times 1, 144\\3 \times 3, 144\\ 1 \times 1, 32 \end{bmatrix} \times 3$    \\ [1em]
\hline 
bottleneck4\_x     & 14$\times$14$\times$64   &  $\begin{bmatrix} 1 \times 1, 192\\3 \times 3, 192\\ 1 \times 1, 64 \end{bmatrix} \times 4$    \\ [1em]
\hline
bottleneck5\_x     & 14$\times$14$\times$96   &  $\begin{bmatrix} 1 \times 1, 384\\3 \times 3, 384\\ 1 \times 1, 96 \end{bmatrix} \times 3$    \\ [1em]
\hline
bottleneck6\_x     & 7$\times$7$\times$160   &  $\begin{bmatrix} 1 \times 1, 576\\3 \times 3, 576\\ 1 \times 1, 160 \end{bmatrix} \times 3$    \\ [1em]
\hline
bottleneck7\_x     & 7$\times$7$\times$320   &  $\begin{bmatrix} 1 \times 1, 960\\3 \times 3, 960\\ 1 \times 1, 320 \end{bmatrix} \times 1$    \\ [1em]
\hline
bottleneck8\_x     & 7$\times$7$\times$1280   &  $ 1 \times 1, 1280$    \\
\hline
encoding / pooling & 8 x 1280 /  1280          &   8 codewords / ave pool   \\ 
\hline 
fc1\_1 / fc1\_2 & 64 / 64  & 10240$\times$64 / 1280$\times$64   \\ 
\hline 
bilinear mapping & 4096  &  -   \\ 
\hline 
fc2 & 128 & 4096$\times$128   \\ 
\hline
classification & n classes & 128$\times$n\\
\hline

\end{tabular}
}
\caption{The architecture of the Deep Encoding Pooling Network based on MobileNet V2 \cite{sandler2018mobilenetv2}. The input image size is $224 \times 224$.}
%\vspace{-0.2in}
\label{table:structure}
\end{table}

\begin{figure*}[h]
\centering
\includegraphics[width= \linewidth]{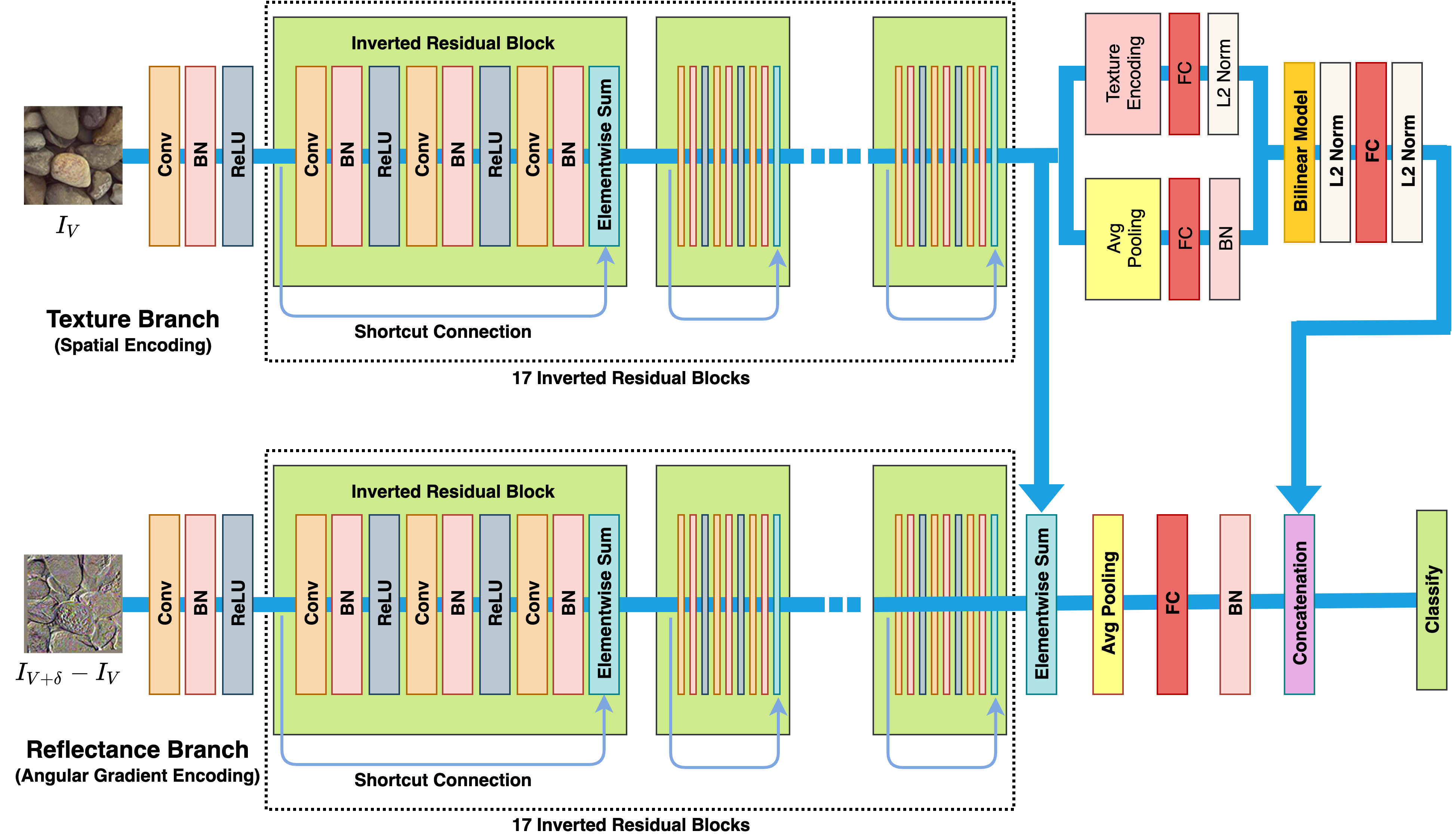}
\caption{A Texture Encoded Angular Network (TEAN) for material recognition. The input to the reflectance branch is the differential angular image, which  captures  material reflectance information via angular gradients. The input to the texture branch is the RGB color image,  to provide the ordered and orderless  spatial information. For the texture branch, we utilize  DEP to balance the orderless texture component and ordered spatial information. The overall architecture of TEAN enables material classification using angular reflectance information, orderless texture and ordered spatial structure.  }
\label{fig:daen}
\end{figure*}

\subsection{Texture Encoded Angular Network (TEAN)}
Adapting the DEP to the RGB image branch in DAIN, we introduce the Texture Encoded Angular Network (TEAN). The detailed network is shown in Figure~\ref{fig:daen}. We develop a two-stream convolutional neural network, one branch input is the differential angular image, representing the material reflectance information. The other branch input is the RGB image, representing the orderless texture details and ordered spatial information. For the color image branch, we utilize the Deep Encoding Pooling Network (DEP) to balance the orderless texture component and ordered spatial information. As in DAIN, we combine feature maps at both intermediate layer and final prediction layer. With the proposed Texture Encoded Angular Network (TEAN), we take advantage of material reflectance information, orderless texture details and ordered spatial information for ground terrain material recognition. This combination of angular cues, orderless spatial cues and ordered spatial cues leads to improved recognition results. 

As shown in Figure~\ref{fig:daen}, we employ ImageNet\cite{deng2009imagenet} pre-trained MobileNet V2\cite{sandler2018mobilenetv2} as the initial prediction unit. As in single view DAIN (Sum), we combine feature maps from the bottlenect8\_x with element-wise sum as intermediate layer combination. Feature maps from color images are fed into the texture encoding layer and the global average pooling layer jointly, followed by bilinear model and fully connected layer, the output from fully connected layer is a 128-D vector. The element-wise summed feature maps are fed into a fully connected layer for dimension reduction, the output is also a 128-D vector. These two 128-D vectors are concatenated and fed into classify layer for material classification. 

\section{Experiments}
\label{sec:experiments}
In this section, we evaluate the performance of DAIN, DEP and TEAN framework for material recognition. 
First, in section~\ref{DAIN_result}, we evaluate which structure of the two stream networks from Figure~\ref{fig:fuse_cmp} works best on the GTOS dataset, leading to the choice in (c) as the DAIN architecture. Based on (c), we consider recognition performance with different DAIN variations for recognition and compare three other state-of-the-art approaches on our GTOS dataset, concluding that multiview DAIN works best.
Second, in section~\ref{DEP_result}, we compare the recognition performance of DEP with fine-tuning MobileNet, bilinear CNN and Deep-TEN. To prove the superior performance of the proposed DEP in material recognition, we experiment DEP on two other material/texture recognition datasets.
Third, in section~\ref{TEAN_result}, we evaluate the performance of TEAN and verify the performance of multi-scale training.
Finally, in section~\ref{conf_matrix}, to gain insight into the performance, we construct the confusion matrix and visualize the features before classification layers with BarnesHut t-SNE \cite{van2014accelerating} for MobileNet, DEP, DAIN and TEAN.

\begin{table*}[h]
\centering
% \resizebox{\linewidth}{!}{%
\begin{tabular}{|l|l|l|l|l|}
\hline 

Model & Method & First input & Second input & Accuracy\\

\hline
\rule{0pt}{0.8\normalbaselineskip}
\multirow{9}{*}{VGG-M \cite{chatfield2014return}} & single view CNN &${I}_{v}$&-&74.3\tiny$\pm2.8$\\
 
& multiview CNN, voting&${I}_{v}$&-&78.1\tiny$\pm2.4$\\

& multiview CNN,3D filter&${I}_{v}$&-&74.8\tiny$\pm3.2$\\
\cline{2-5}
\rule{0pt}{0.8\normalbaselineskip}
& single view DAIN (Sum)&${I}_{v}$&${I}_{v+\delta}$&77.5\tiny $\pm2.7$\\
 
& single view DAIN (Sum)&${I}_{v}$&${I}_{\delta}$&79.4\tiny $\pm3.4$\\

& single view DAIN (Max)&${I}_{v}$&${I}_{\delta}$&79.0\tiny $\pm1.8$\\

\cline{2-5}
\rule{0pt}{0.8\normalbaselineskip}
& multiview (Sum/voting)&${I}_{v}$&${I}_{\delta}$&80.0\tiny $\pm2.1$\\

& multiview DAIN (Sum/pooling)&${I}_{v}$&${I}_{\delta}$&81.2\tiny $\pm1.7$\\

& multiview DAIN (3D filter/pooling)&${I}_{v}$&${I}_{\delta}$&81.1\tiny $\pm1.5$\\
\hline 
\rule{0pt}{0.8\normalbaselineskip}
\multirow{5}{*}{MobileNet V2 \cite{sandler2018mobilenetv2}} & single view CNN &${I}_{v}$&-&80.4\tiny$\pm3.2$\\
 
& multiview CNN, voting&${I}_{v}$&-&82.5\tiny$\pm2.8$\\

\cline{2-5}
\rule{0pt}{0.8\normalbaselineskip}
 
& single view DAIN (Sum)&${I}_{v}$&${I}_{\delta}$&82.5\tiny $\pm2.3$\\

& multiview DAIN (Sum/voting)&${I}_{v}$&${I}_{\delta}$&85.8\tiny $\pm2.6$\\

& multiview DAIN (Sum/pooling)&${I}_{v}$&${I}_{\delta}$&86.2\tiny $\pm2.5$\\      
\hline 

\end{tabular}
% }
\caption{Results comparing performance of standard CNN recognition without angular differential imaging (first three rows) to our single-view DAIN (middle three rows) and our multi-view DAIN (bottom three rows). ${I}_{v}$ denotes the image from viewpoint $v$, $I_{v+\delta}$ is the image obtained from viewpoint $v+\delta$, and $I_\delta=I_v - {I}_{v+\delta}$ is the differential image. The differential angular imaging network (DAIN) has superior performance over CNN even when comparing single view DAIN to multiview CNN.
Multiview DAIN provides the best recognition rates.}
\label{table:method_comparison}
\end{table*}

% The first evaluation determines which structure of the two stream networks from Figure~\ref{fig:fuse_cmp} works best on the GTOS dataset, leading to the choice in (c) as the DAIN architecture. 
% The second evaluation considers recognition performance with different DAIN variations for recognition. 
% The third experimental evaluation compares three other state-of-the-art approaches on our GTOS-dataset, concluding that multiview DAIN works best. 
% The 4th experimental evaluation proves that comparing with CNN structures designed for object classification, the proposed DEP generates superior performance in material recognition. 
% Finally, Adapting the DEP to the RGB image branch in DAIN, we introduce Texture Encoded Angular Network (TEAN) and evaluate the performance of the new framework. 
% To gain insight into the performance, we construct the confusion matrix and visualize the features before classification layers with BarnesHut t-SNE \cite{van2014accelerating} for MobileNet, DEP, DAIN and TEAN.  

\vspace{.1in}
\noindent\textbf{Training procedure}
We design 5 training and testing splits by assigning about 70\% of ground terrain surfaces of each class to training and the remaining 30\% to testing.
In order to ensure that there is no overlap between training and testing sets, if one sample is in the training set, all views and illumination conditions for that sample is in the training set. 
Each input image from our GTOS database is resized into 240 $\times$ 240.
Since the snow class only has 2 samples in the dataset, we omit this class from experiments.

Comparing with recent mobile platform designed MobileNet V2 \cite{sandler2018mobilenetv2}, the number of parameters for VGG-M \cite{chatfield2014return} is tremendous. So for training a VGG-M based two branch network, we first fine-tune the VGG-M model separately with RGB and differential images with batch size 196, dropout rate 0.5, momentum 0.9. 
We employ the augmentation method that horizontally and vertically stretch training images within $\pm10\%$, with an optional 50\% horizontal and vertial mirror flips.
The images are randomly cropped into 224 $\times$ 224 material patches.
All images are pre-processed by subtracting a per color channel mean and normalizing for unit variance.
The learning rate for the last fully connected layer is set to 10 times of other layers.
We first fine-tune only the last fully connected layer with learning rate $5 \times {10}^{-2}$ for 5 epochs;
then, fine-tune all the fully connected layers with learning rate ${10}^{-2}$ for 5 epochs.
Finally we fine-tune all the layers with leaning rate starting at ${10}^{-3}$, and decrease by a factor of 0.1 when the training accuracy saturates.
For MobileNet V2 \cite{sandler2018mobilenetv2} based DAIN, we train the network end-to-end on GTOS dataset directly.

Following prior works \cite{xue2016differential, xue2018deep}, for the fine-tuned two-branch VGG-M model and two-branch MobileNet V2 model, we experiment with batch size 64 and learning rate starting from 0.01 which is reduced by a factor of 0.1 when the training accuracy saturates.
We augment training data with randomly stretch training images by $\pm25\%$ horizontally and vertically, and also horizontal and vertical mirror flips with 50\% chance.
The images are randomly cropped into 224 $\times$ 224 material patches.
We first backpropagate only to feature maps combination layer for 3 epochs, then fine tunes all layers.
We employ the same augmentation method for the multiview images of each material surface.
We randomly select the first viewpoint image, then subsequent $N=4$ view point images are selected for experiments.

\begin{table}[h]
\centering
\begin{tabular}{|l|M{1.5cm}|M{2cm}|M{1.5cm}|l|}
\hline 
Method & Final Layer Combination  & Intermediate Layer Combination  & Intermediate and Final layer Combination \\
\hline 
Accuracy&77.0\tiny $\pm2.5$&74.8\tiny $\pm3.4$&79.4\tiny $\pm3.4$ \\
\hline 
\end{tabular}
\caption{Comparison of accuracy from different two stream methods as shown in Figure~\ref{fig:fuse_cmp}. The feature-map combination method for (b) and (c) is Sum at Conv5 layers after ReLU. The reported result is the mean accuracy and the subscript shows the standard deviation over 5 splits of the data. Notice that the architecture in (c) gives the best performance and is chosen for the network architecture.}
\label{table:sum_comparison}
\end{table}

\subsection{DAIN Recognition Results}
\vspace{.1in}

\noindent\textbf{Recognition Benchmarks}
To evaluate the recognition performance of DAIN, we employ the ImageNet \cite{deng2009imagenet} pre-trained VGG-M model \cite{chatfield2014return} as the initial prediction unit. We compare DAIN with both single view and multiview CNNs. As in single view CNN, we follow the standard procedure to fine-tune pre-trained networks, by replacing the classification layer with a new 39-way classification layer. Since for our GTOS database, each sample is observed with multiple viewing angles, we set multiview CNN baseline to demonstrate the effectiveness of multiview observation. To incorporate multiview information, for baseline comparison, we use two different methods:
(1) voting: use the predictions from each view to vote
(2) 3D filter: follow \cite{tran2015learning} to use a $3\times 3 \times 3$ learned filter bank to convolve the multiview feature maps.

\vspace{.1in}
\noindent\textbf{Evaluation for DAIN Architecture} Table~\ref{table:sum_comparison} shows the mean classification accuracy of the different three branch combination methods depicted in Figure~\ref{fig:fuse_cmp}. 
Inputs are single view images (${I}_{v}$) and single view differential images (${I}_{\delta}$).
Combining the two streams at the final prediction layer  (77\% accuracy) is compared with the intermediate layer combination  (74.8\%) or the hybrid approach in Figure~\ref{fig:fuse_cmp} (c) (79.4\%) which we choose as the network architecture for following experiments. 
The combination method used is Sum and the feature maps are obtained from  Conv5 layers after ReLU.

\vspace{.1in}
\noindent\textbf{DAIN Recognition Performance}\label{DAIN_result}
We evaluate DAIN recognition performance for single view input (and differential image) and for multiview input from the GTOS database. Additionally, we compare the results to recognition using a standard CNN without a differential image stream. For all multiview experimental results we choose the number of viewpoints $N=4$, separated by $10^\circ$ with the starting viewpoint chosen at random (and the corresponding differential input). 
Table~\ref{table:method_comparison} shows the resulting recognition rates (with standard deviation over 5 splits shown as a subscript).  
The first three rows shows the accuracy {\it without} differential angular imaging, using both single view and multiview input. Notice the recognition performance for these non-DAIN results are generally lower than the DAIN recognition rates in the rest of the table. 
The middle three rows show the recognition results for single view DAIN.
For combining feature maps we evaluate both Sum and Max which have comparable results. 
Notice that single view DAIN achieves better recognition accuracy than multiview CNN with voting (79.4\% vs. 78.1\%). This is an important result indicating the power of using the differential image. Instead of four viewpoints separated by $10^\circ$, a single viewpoint and its differential image achieves a better recognition. These results provide design cues for  building imaging systems tailored to material recognition.
We also evaluate whether using inputs from the two viewpoints directly (i.e. $I_v$ and $I_{v+\delta}$) is comparable to using $I_v$ and the differential image $I_{\delta}$.
Interestingly, the differential image as input has an advantage (79.4\% over 77.5\%).
The last three rows for VGG-M model of Table~\ref{table:method_comparison} 
show that recognition performance using multiview DAIN beats the performance of both single view DAIN and CNN methods with no differential image stream.
We evaluate different ways to combine the multiview image set including voting, pooling, and the 3D filter+pooling illustrated in Figure~\ref{fig:3D_filter}.

\begin{table}
\centering

\begin{tabular}{|l|l|}
\hline 

 Architecture & Accuracy\\

\hline 
FV+CNN\cite{cimpoi2014describing}&75.4\%\\%corrected
 
FV-N+CNN+N\tiny3D \normalsize\cite{degol2016geometry}&58.3\%\\%corrected

MVCNN\cite{su2015multi}&78.1\%\\%corrected

\textbf{multiview DAIN (Sum/pooling)}&\textbf{81.2\%}\\%corrected
         
\hline 
\end{tabular}
\caption{Comparison with the state of art algorithms on GTOS dataset. Notice that our method, multiview DAIN, achieves the best recognition accuracy. 
}
\label{table:state_art_comparison}
\end{table}

Table~\ref{table:state_art_comparison} shows the recognition rates for
multiview DAIN that outperforms three other multi-view classification method:
FV+CNN\cite{cimpoi2014describing}, FV-N+CNN+N\tiny3D \normalsize\cite{degol2016geometry}, and MVCNN\cite{su2015multi}.
The table shows recognition rates for a single split of the GTOS database 
with images resized to 240 $\times$ 240.
All experiments are based on the same pre-trained VGG-M model.
We use the same fine-tuning and training procedure as in the MVCNN\cite{su2015multi} experiment.
For FV-N+CNN+N\tiny3D \normalsize applied to GTOS, 10 samples (out of 606) failed to get geometry information by the method provided in \cite{degol2016geometry} and we removed these samples from the experiment. 
The patch size in \cite{degol2016geometry} is 100 $\times$ 100, but the accuracy for this patch size for GTOS was only 43\%, so we use 240 $\times$ 240. 
We implement FV-N+CNN+N\tiny3D \normalsize with linear mapping instead of homogeneous kernel map\cite{vedaldi2012efficient} for SVM training to save memory with this larger patch size.

\vspace{.1in}
\noindent\textbf{DAIN with MobileNet V2}
The CNN module of the two stream network can be replaced by other state-of-the-art deep learning methods to further improve results. To demonstrate this, we change the CNN module to ImageNet pre-trained MobileNet V2 \cite{sandler2018mobilenetv2}. Combining feature maps generated from the bottleneck8\_x (the eighth Bottleneck inverted residual block) with training batch size 64. The recognition results are shown in Table~\ref{table:method_comparison}. After replacing VGG-M model with MobileNet V2, the recognition performance improves for all the methods. But still our DAIN network architecture performs better than the non-DAIN methods. Notice that sngle view DAIN achieves same recognition accuracy with smaller variance than multiview CNN with voting. The best recognition performance is the multiview DAIN (Sum/pooling) with single view images (${I}_{v}$) and single view differential images (${I}_{\delta}$) as inputs, which is 86.2\%.

\begin{figure}[t]
\centering
\includegraphics[width= .95\linewidth]{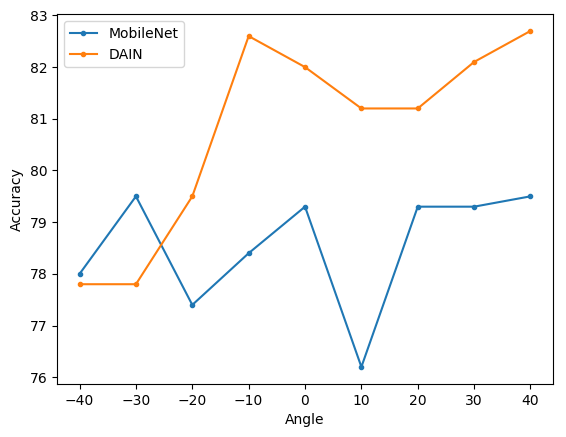}
\caption{The recognition accuracy of fine-tuned MobileNet V2 and DAIN on different observation angles. The recognition accuracy of DAIN outperforms fine-tuned MobileNet V2 in different observation angles.}
\label{fig:angle_performance}
\end{figure}

\begin{table*}[h]
\centering
\begin{tabular}{|l|c|c|c|c|c|c|c|c|c|c|}
\hline
Material Class & painted asphalt & brick & \textbf{cement} & \textbf{dry grass} & \textbf{limestone} & moss & mud & painted turf & \textbf{sand} & stone \\ \hline
DAIN & 97.7 & 90.8 & 89.9 & 57.8 & 93.9 & 76.8 & 93.5 & 97.3 & 84.2 & 71.6 \\ \hline
MobileNet V2 & 98.1 & 97.3 & 72.3 & 18.5 & 87.5 & 91.2 & 98.1 & 98.9 & 56.4 & 68.1 \\ \hline
\end{tabular}
\caption{The recognition accuracy of MobileNet V2 based DAIN and fine-tuned MobileNet V2 on different material classes.}
\label{table:material}
\end{table*}

\begin{figure}[t]
\centering
\subfloat[painted asphalt]
{
  \includegraphics[width=.23\linewidth]{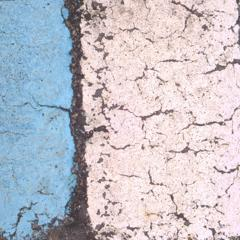}
}
\subfloat[brick]
{
  \includegraphics[width=.23\linewidth]{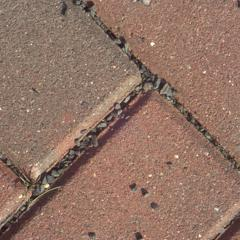}
}
\subfloat[painted turf]
{
  \includegraphics[width=.23\linewidth]{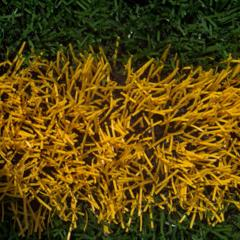}
  
}
\subfloat[stone]
{
  \includegraphics[width=.23\linewidth]{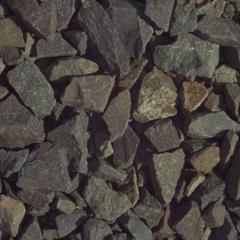}
}\\
\subfloat[dry grass]
{
  \includegraphics[width=.23\linewidth]{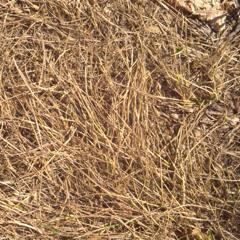}
}
\subfloat[limestone]
{
  \includegraphics[width=.23\linewidth]{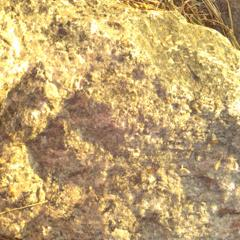}
}
\subfloat[cement]
{
  \includegraphics[width=.23\linewidth]{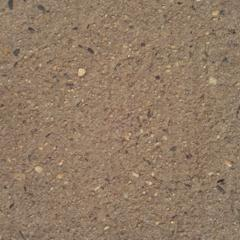}
  
}
\subfloat[sand]
{
  \includegraphics[width=.23\linewidth]{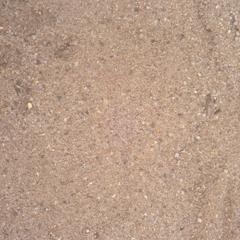}
}
\caption{The sample images for different material classes. 
For materials with distinct shape or color information like painted asphalt, brick, painted turf and stone (top), the recognition performance is similar. But for material classes where recognition depends on  material reflectance and  fine-scale texture  (like cement, dry grass, limestone and sand), angular gradients are an important cue and DAIN significantly outperforms MobileNet V2.
}
\label{fig:material_classes}
\end{figure}

\vspace{.1in}
\noindent\textbf{Differential Angular Imaging Analysis}
To further analyze the effect of differential angular images for recognition performance, based on GTOS split 1, we compare the recognition difference of fine-tuned MobileNet V2 and MobileNet V2 based DAIN for different observation angles and different material classes. For different observation angles, we compare the test set recognition accuracy. The result is shown in Figure~\ref{fig:angle_performance}; notice the recognition accuracy increases when observation angle moves to the center (0 degree). Also with the help of differential angular images, the recognition accuracy of DAIN outperforms fine-tuned MobileNet V2 in different observation angles. 
We select 10 classes from the GTOS dataset to compare the test set recognition accuracy for different material classes. The results are shown in Table~\ref{table:material}. For materials with distinct shape or color information like painted asphalt, brick, painted turf and stone, the recognition performance is similar. But for material classes where recognition depends on  material reflectance and  fine-scale texture  (like cement, dry grass, limestone and sand, see Figure~\ref{fig:material_classes}), angular gradients are an important cue and DAIN significantly outperforms MobileNet V2.

\subsection{DEP Recognition Results} \label{DEP_result}
\vspace{.1in}

\noindent\textbf{Recognition Benchmarks}
We compare the DEP network with the following three baseline methods based on ImageNet \cite{deng2009imagenet} pre-trained MobileNet V2 \cite{sandler2018mobilenetv2}: (1) CNN with global average pooling (MobileNet), (2) CNN with texture encoding (Deep-Ten) and (3) CNN with bilinear models (Bilinear-CNN).
All three methods support end-to-end training. For equal comparison, we employ the same training procedure aforementioned, and we use an identical training and evaluation procedure for each experiment.

CNN with global average pooling (MobileNet): As in single view CNN, we follow the standard procedure to fine-tune pre-trained MobileNet, by replacing the classification layer with a new 39-way classfication layer. 
The global average pooling works as feature pooling that encodes the 7$\times$7$\times$1280 dimensional features from the pre-trained MobileNet V2 into a 1280 dimensional vector.

CNN with texture encoding (Deep-TEN):
The Deep Texture Encoding Network (Deep-TEN) \cite{zhang2016deep} embeds the texture encoding layer on top of the 50-layer pre-trained ResNet \cite{he2016deep}. To make an equal comparison, we replace the 50-layer ResNet with MobileNet V2.
As in \cite{zhang2016deep}, we reduce the number of CNN streams outputs channels from 1280 to 128 with a 1$\times$1 convolutional layer. We replace the global average pooling layer in the MobileNet V2 with texture encoding layer, set the number of codewords to 32 for experiments. Outputs from the texture encoding layer are normalized with L2 normalization. A fully connected layer with soft max loss follows the texture encoding layer for classification.

CNN with bilinear models (Bilinear-CNN):
Bilinear-CNN \cite{lin2015bilinear}  employs bilinear models with feature maps from convolutional layers. Outputs from convolutional layers of two CNN streams are multiplied using outer product at each location and pooled for recognition. To make an equal comparison, we employ the pre-trained MobileNet V2 as CNN streams for feature extractor. Feature maps from the last convolutional layer are pooled with bilinear models. we reduce the number of CNN streams outputs channels from 1280 to 128 with a 1$\times$1 convolutional layer before bilinear models. The dimension of feature maps for bilinear models is 7$\times$7$\times$128 and the pooled bilinear feature is of size 128$\times$128. The pooled bilinear feature is fed into classification layer for classification.

\vspace{.1in}
\noindent\textbf{DEP Recognition Performance}
Table \ref{table:gtos} is the classification accuracy of fine-tuning MobileNet V2 ~\cite{sandler2018mobilenetv2}, Bilinear CNN~\cite{lin2015bilinear}, Deep-TEN~\cite{zhang2016deep} and the proposed DEP on the GTOS dataset. The recognition accuracy for combining spatial information and texture details (DEP) is 83.3\%. That's 2.5\% better than only focusing on spatial information (ResNet) and 2.5\% better than only focusing on texture details (Deep-TEN).

\begin{table}[t]
\centering
\begin{tabular}{|l|l|l|l|}
\hline
MobileNet \cite{sandler2018mobilenetv2} & Bilinear CNN\cite{lin2015bilinear} &	Deep-TEN\cite{zhang2016deep} & DEP ({\small\TextRed{ours}}) \\ \hline
 80.4\tiny$\pm3.2$ & 80.8\tiny$\pm2.2$  & 80.8\tiny$\pm1.5$ &\textbf{83.3\tiny$\pm2.1$} \\ \hline
\end{tabular}
\caption{Comparison our Deep Encoding Pooling Network (DEP) with MobileNet V2 (left) \cite{sandler2018mobilenetv2}, Bilinear CNN (mid) \cite{lin2015bilinear} and Deep-TEN (right) \cite{zhang2016deep} on GTOS dataset. For MobileNet, we replace the 1000-way classification layer with a new classification layer, the output dimension of new classification layer is the number of classes, which is 39 for GTOS.}
\label{table:gtos}
\end{table}

\vspace{.1in}
\noindent\textbf{Evaluation on MINC and DTD Dataset} To show the generality of DEP for material recognition, we experiment on two other material/texture recognition datasets: Describable Textures Database (DTD) \cite{cimpoi2014describing} and Materials in Context Database (MINC) \cite{bell2015material}. For an equal comparison, we build DEP based on a 50-layer ResNet \cite{he2016deep}, the feature maps channels from CNN streams are reduced from 2048 to 512 with a 1$\times$1 convolutional layer. 
The result is shown in Table~\ref{table:state-art}, DEP outperforms the state-of-the-art on both datasets. 
Note that we only experiment with single scale training. As mentioned in \cite{lin2015bilinear}, multi-scale training is likely to improve results for all methods.

\begin{table}[t]
\centering
\begin{tabular}{|c|c|c|}
\hline
Method & DTD\cite{cimpoi2014describing}    & Minc-2500\cite{bell2015material} \\ \hline
FV-CNN~\cite{cimpoi2015deep} & 72.3\% & 63.1\%    \\ \hline
Deep-TEN~\cite{zhang2016deep} & 69.6\% & 80.4\%    \\ \hline
DEP ({\small\TextRed{ours}})    & \textbf{73.2}\% & \textbf{82.0}\%    \\ \hline
\end{tabular}
\caption{Comparison with state-of-the-art algorithms on Describable Textures Dataset (DTD) and Materials in Context Database (MINC).}
\label{table:state-art}
%\vspace{-0.2in}
\end{table}

\begin{figure*}[t]
\centering
\subfloat
{
\includegraphics[width=.23\linewidth]{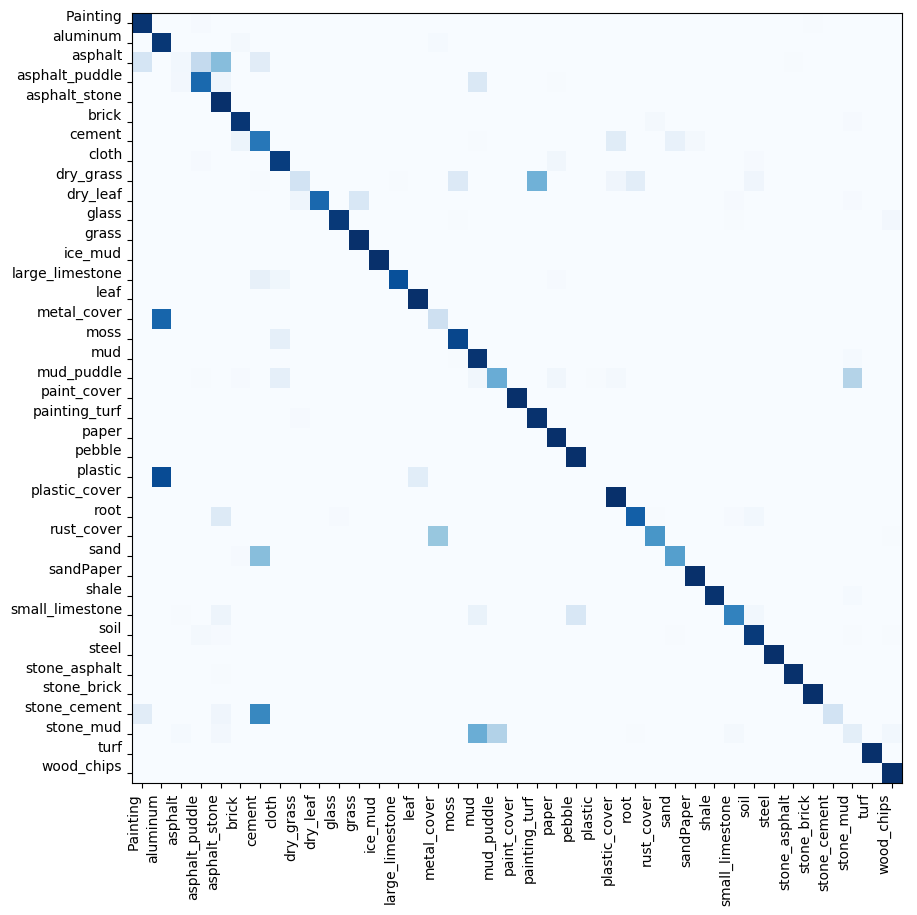}
}
\subfloat
{
\includegraphics[width=.23\linewidth]{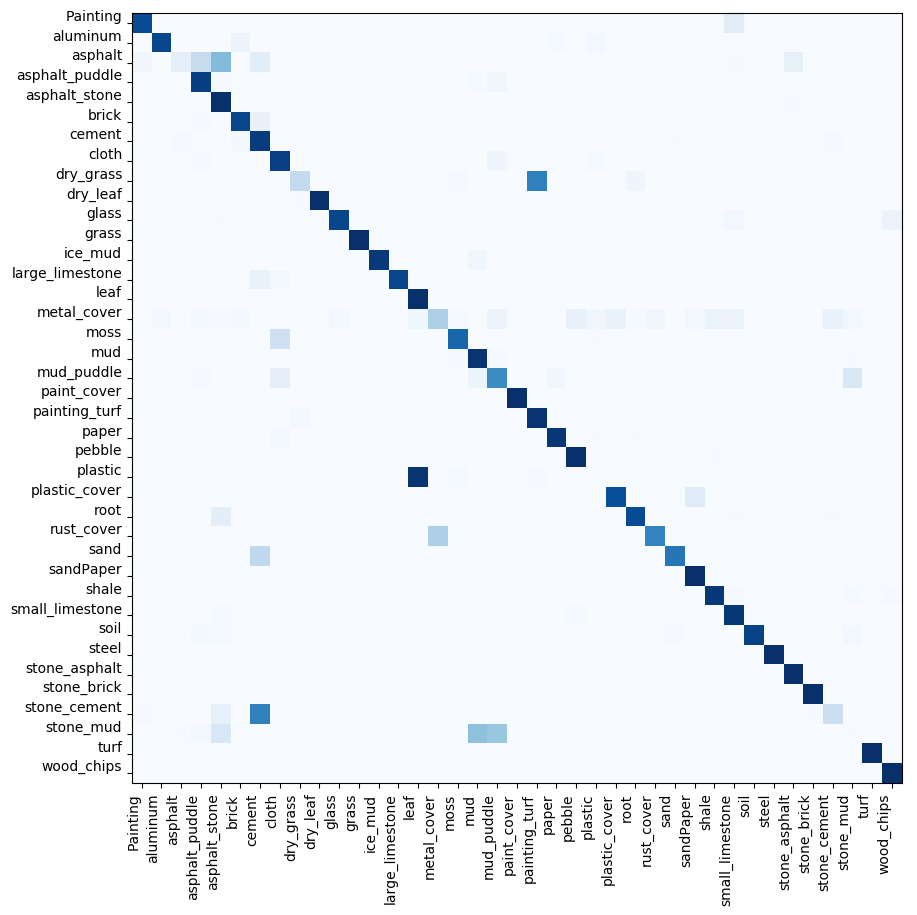}
}
\subfloat
{
\includegraphics[width=.23\linewidth]{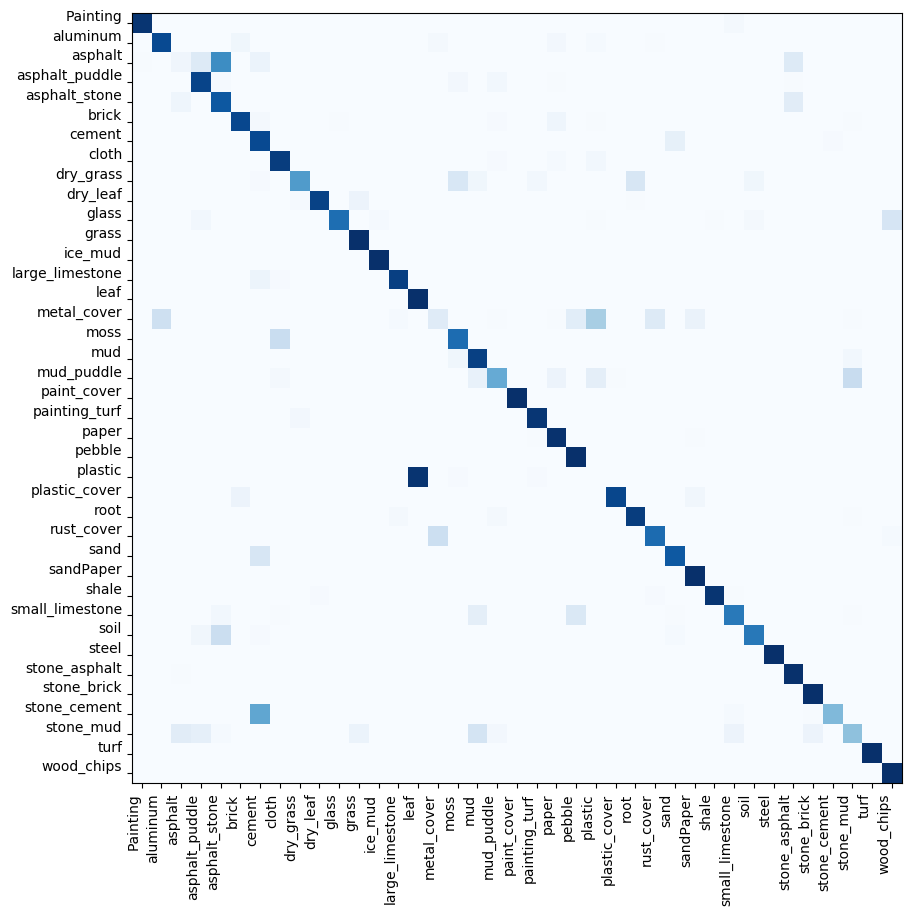}
}
\subfloat
{
\includegraphics[width=.23\linewidth]{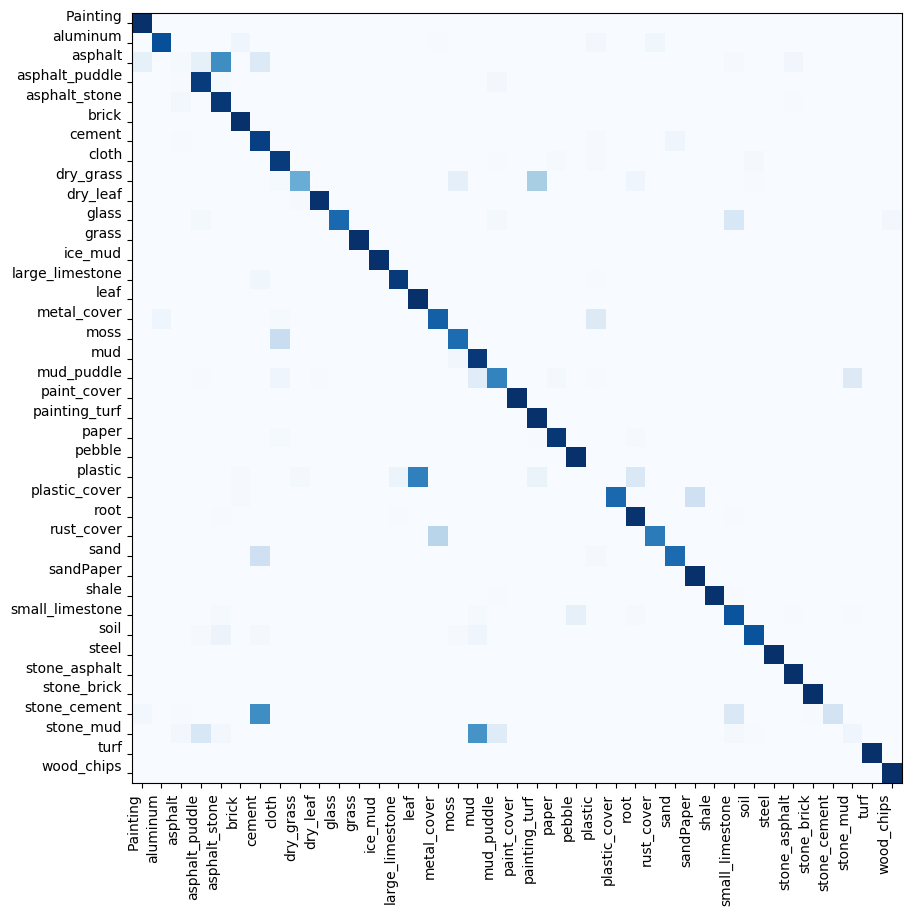}
}
\setcounter{subfigure}{0}
\subfloat[mobilenet]
{
\includegraphics[width=.23\linewidth]{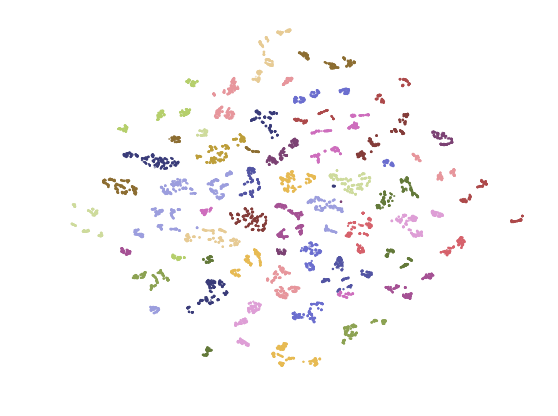}
}
\subfloat[DEP]
{
\includegraphics[width=.23\linewidth]{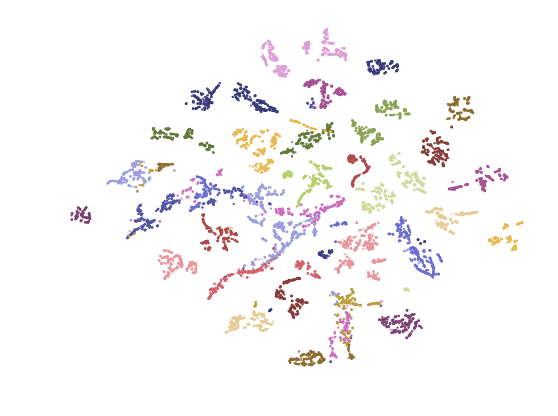}
}
\subfloat[DAIN]
{
\includegraphics[width=.23\linewidth]{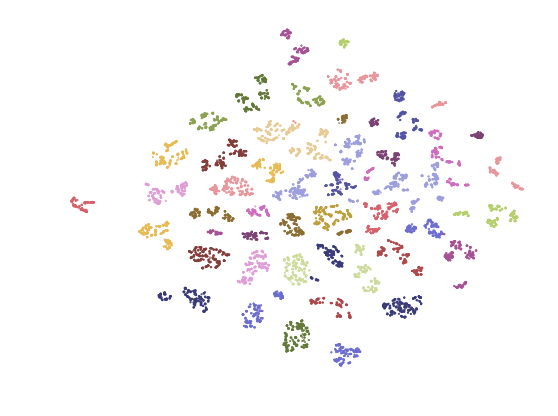}
}
\subfloat[TEAN]
{
\includegraphics[width=.23\linewidth]{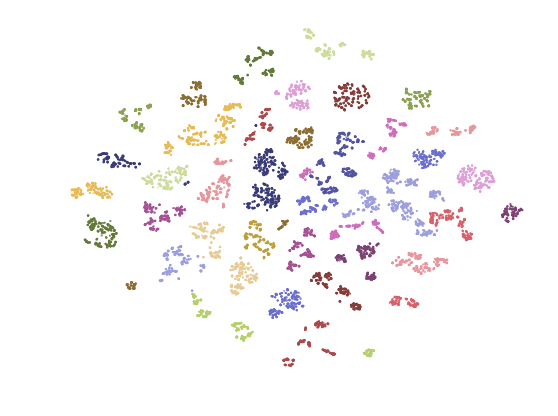}
}
\caption{The Barnes-Hut t-SNE \cite{van2014accelerating} and confusion matrix of four material recognition models based on GTOS : MobileNet (left), DEP (mid left), DAIN (mid right) and TEAN (right). For Barnes-Hut t-SNE \cite{van2014accelerating},  we employ images from validation set and extract features before classification layers of four models for experiment. We see that TEAN separates and clusters the classes better. (Dark blue represents higher values and light blue represents lower values in the confusion matrix.)}
\label{fig:tsne}
\end{figure*}

\subsection{TEAN Recognition Results} \label{TEAN_result}
% \vspace{.1in}
% \noindent\textbf{TEAN Recognition Performance}
Table~\ref{table:dain_daen} is the mean classification accuracy comparison of MobileNet V2 based single view/multiview CNN fine-tune, DEP, DAIN and TEAN. As in DAIN, we experiment with voting and pooling to combine the multiview image set. From the result we can see that multiview TEAN (Sum/pooling) performs best, the recognition accuracy is 87.6\%, which is 5.1\% better than multiview CNN, voting baseline. Also the recognition performance for TEAN outperforms DAIN in both single view and multiview.

\begin{table}[t]
\centering
\begin{tabular}{|l|l|}
\hline
Method & Accuracy \\ \hline
single view CNN &80.4\tiny$\pm3.2$\\ 
multiview CNN, voting &82.5\tiny$\pm2.8$\\ \hline
single view DEP &83.3\tiny$\pm2.1$\\
multiview DEP, voting &85.8\tiny$\pm1.9$\\ \hline
single view DAIN (Sum) & 82.5\tiny$\pm2.3$  \\ 
multiview DAIN (Sum/voting) & 85.8\tiny $\pm2.6$ \\ 
multiview DAIN (Sum/pooling)& 86.2\tiny $\pm2.5$ \\ \hline
single view TEAN (Sum) & 84.7\tiny$\pm2.1$  \\ 
multiview TEAN (Sum/voting) & 87.4\tiny $\pm2.3$ \\ 
multiview TEAN (Sum/pooling)& 87.6\tiny $\pm2.0$ \\ \hline
\end{tabular}
\caption{Results comparing performance of CNN fine-tune, DEP, DAIN and TEAN based on MobileNet V2 \cite{sandler2018mobilenetv2}.}
\label{table:dain_daen}
\end{table}

\vspace{.1in}
\noindent\textbf{Multi-scale Training}
Multi-scale training is a common image augmentation trick to simulate observing materials at different distances \cite{degol2016geometry,zhang2016deep,xue2018deep}. We also experiment this with our GTOS dataset. We resize images into different resolutions, and randomly crop 224$\times$224 patches for training. Following prior works \cite{zhang2016deep,xue2018deep}, We experiment TEAN with two groups of resolution settings: (256$\times$256, 384$\times$384, 512$\times$512) and (224$\times$224, 246$\times$246, 268$\times$268). For training/testing split 1, the recognition accuracy is 81.93\% and 82.03\% respectively, it is lower than  the single view TEAN, in which the accuracy is 82.87\%. Although the result is contrary with prior works \cite{degol2016geometry,zhang2016deep,xue2018deep}, that simulating observing materials at different distances with multi-scale training is helpful for performance, we think the result is meaningful for GTOS. Since images in the GTOS dataset are captured with a fixed distance between the camera and ground terrain, the observing distance is constant for all the images. We conclude the multi-scale training is not helpful for our GTOS dataset.

\subsection{Confusion Matrix and Feature Visualization} \label{conf_matrix}
% \vspace{.1in}
% \noindent\textbf{Framework Comparison}
To gain insight into why TEAN performs best for material recognition, based on training/testing split 1, we compute the confusion matrix of MobileNet, DEP, DAIN and TEAN and visualize features before classification layers with Barnes-Hut t-SNE \cite{van2014accelerating}. For features visualization, we employ images from validation set and extract features before classification layers of four models for experiment.
The result is shown in Figure \ref{fig:tsne}. Notice that TEAN separates and clusters the classes better.

\section{Conclusion}
In summary, there are three main contributions of this work: 1) The GTOS Dataset with ground terrain imaged by systematic in-scene measurement of partial reflectance instead of in-lab reflectance measurements. The database contains 34,243 images with 40 surface classes, 18 viewing directions, 4 illumination conditions, 3 exposure settings per sample and several instances/samples per class; 2) Differential Angular Imaging for a sparse representation of the spatial distribution of angular gradients that provides key cues for material recognition; 
3) We develop and evaluate architectures for using differential angular imaging, texture details and spatial information for material recognition, showing superior results for differential inputs as compared to original images. 
Our work in measuring and modeling outdoor surfaces has important implications for applications such as robot navigation (determining control parameters based on current ground terrain) and automatic driving (determining road conditions by partial real time reflectance measurements). The database and methods will provides a  foundation for additional in-depth studies of material recognition in the wild.

\ifCLASSOPTIONcompsoc
  \section*{Acknowledgments}
\else
  \section*{Acknowledgment}
\fi

This work was supported by National Science Foundation award IIS-1421134 and IIS-1715195. A GPU used for this research was donated by the NVIDIA Corporation.

\ifCLASSOPTIONcaptionsoff
  \newpage
\fi



%\bibliographystyle{IEEEtran}
%\bibliography{IEEEabrv,../bib/paper}IEEEtran
\bibliographystyle{IEEEtran}
\bibliography{main}


\begin{IEEEbiography}[{\includegraphics[width=1in,height=1.25in,clip,keepaspectratio]{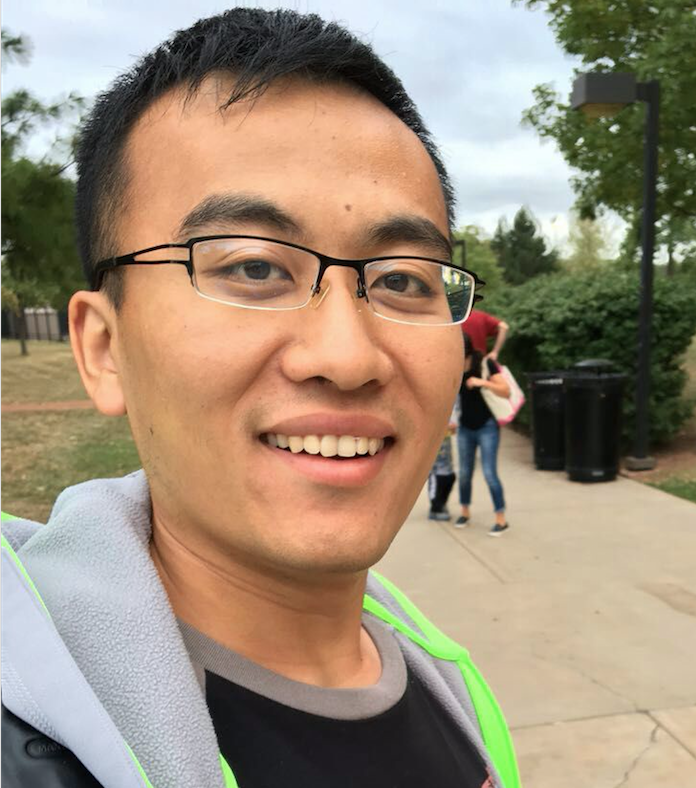}}]{Jia~Xue}
is a Ph.D. student of Electrical Computer Engineering at Rutgers University--New Brunswick, New Brunswick, NJ. He received the bachelor's degree from University of Electronic Science and Technology of China in 2015. His research interests lie in computer vision and machine learning, including material and texture recognition, segmentation, high-level and mid-level vision.
\end{IEEEbiography}

%\begin{IEEEbiography}[{\includegraphics[width=1in,height=1.25in,clip,keepaspectratio]{figure/hang.jpg}}]{Hang Zhang}
\begin{IEEEbiography}[{\includegraphics[width=1in,height=1.25in,clip,keepaspectratio]{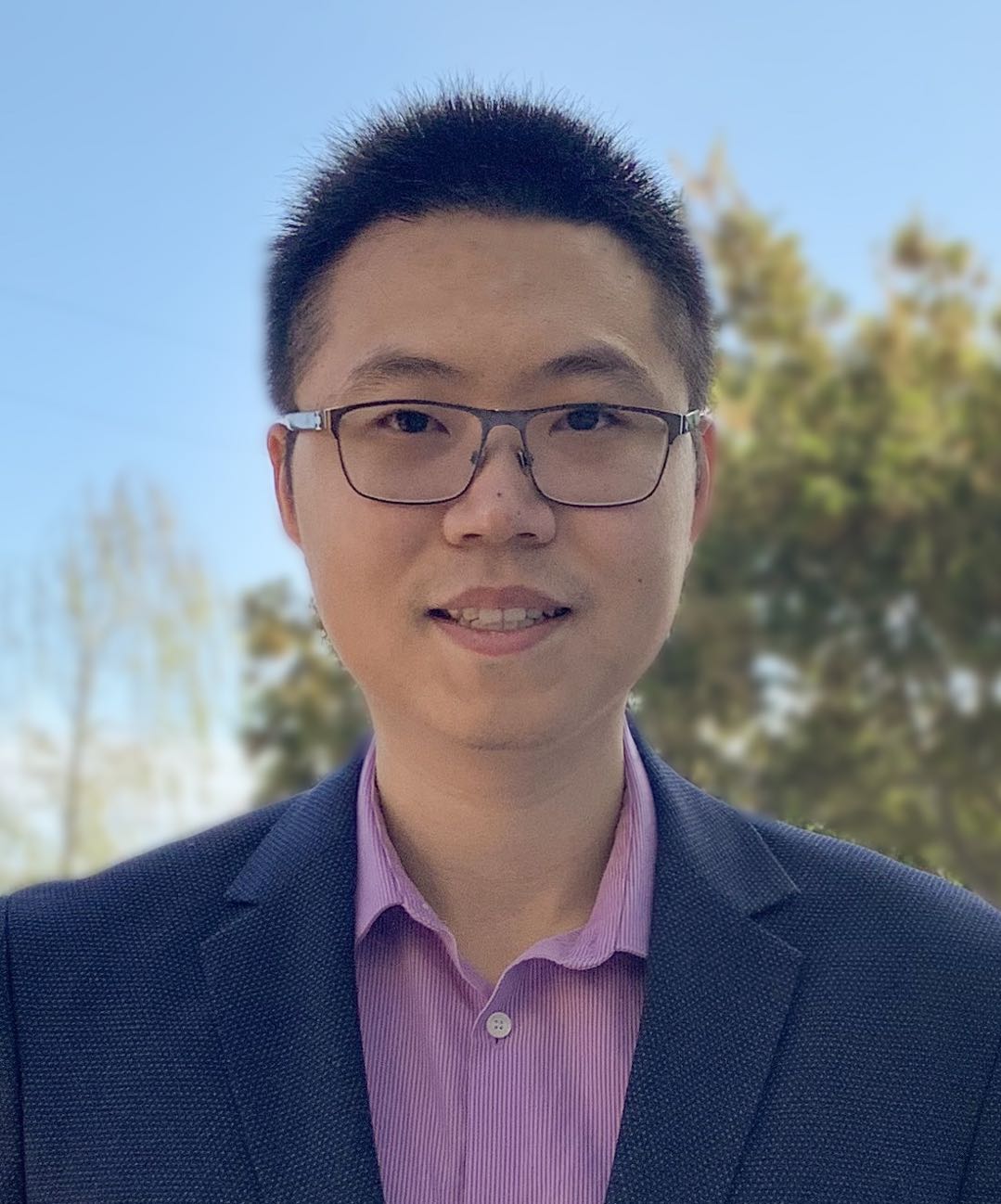}}]{Hang Zhang}
Dr. Hang Zhang is an Applied Scientist working at Amazon Web Service Inc. 
Prior to joining Amazon, he received the PhD degree with Prof. Kristin Dana at Rutgers University in 2017.  
Before coming to Rutgers, he received the BS degree from Southeast University (China) in 2013.   
His research interests are material and texture recognition, neural style transfer, semantic segmentation and large scale image classification. 

\end{IEEEbiography}

\begin{IEEEbiography}[{\includegraphics[width=1in,height=1.25in,clip,keepaspectratio]{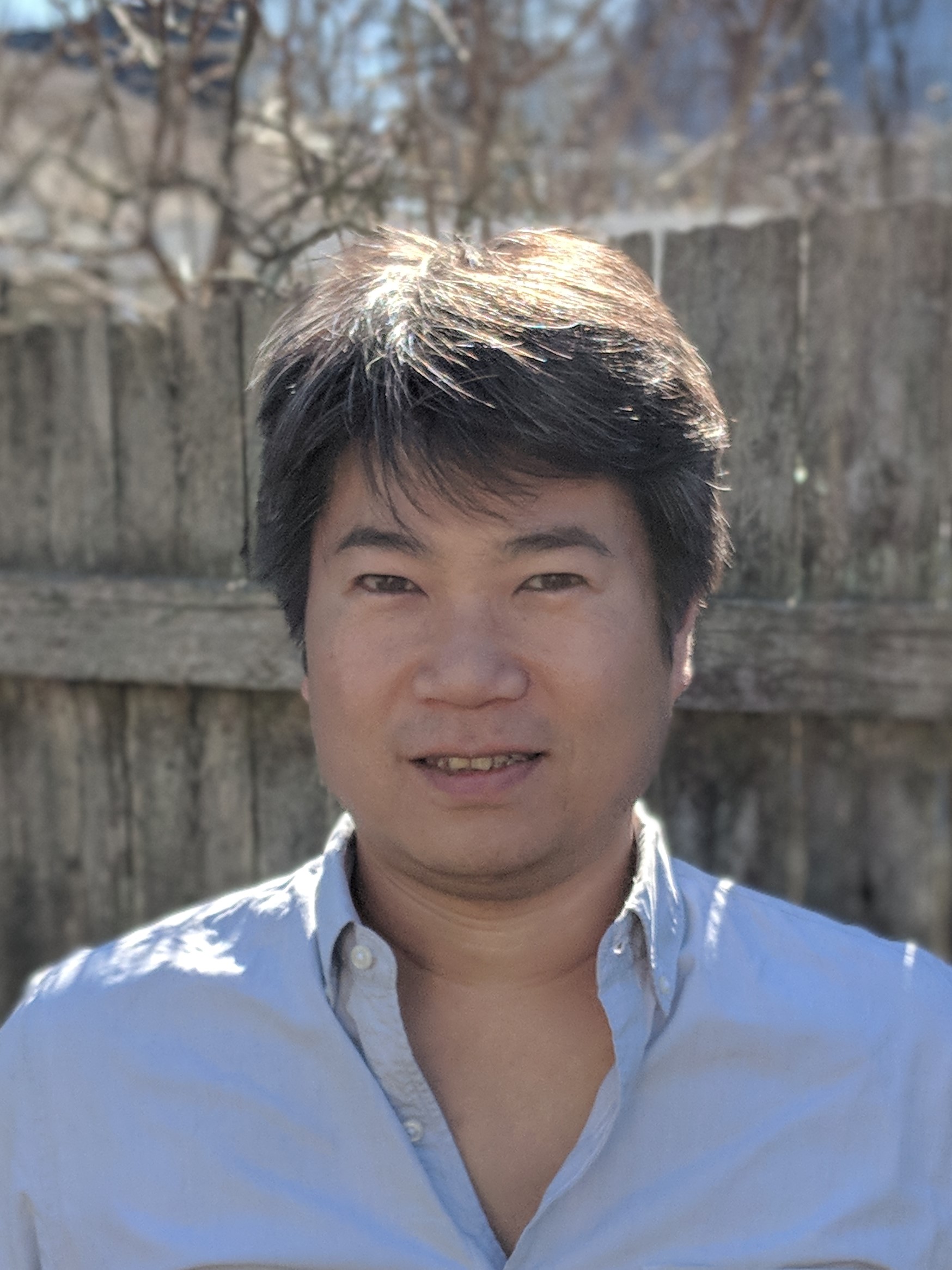}}]{Ko Nishino}
is a Professor in the Department of Intelligence Science and Technology at Kyoto University. He received his B.E. and M.E. in Information and Communication Engineering in 1997 and 1999, respectively, and PhD in Computer Science in 2002, all from University of Tokyo. Before joining Kyoto University in 2018, he was a Professor in the Department of Computer Science at Drexel University. His primary research interests lie in computer vision and machine learning including appearance modeling and material recognition, human behavior analysis, and computational photography. He received the NSF CAREER award in 2008.
\end{IEEEbiography}

\begin{IEEEbiography}[{\includegraphics[width=1in,height=1.25in,clip,keepaspectratio]{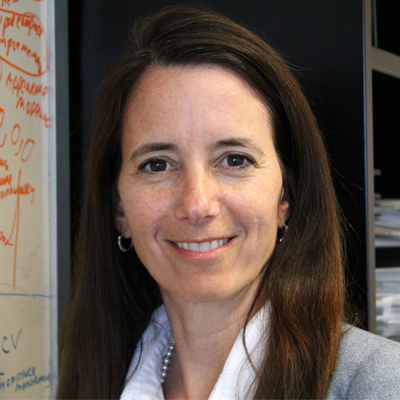}}]{Kristin~J.~Dana}
Dr. Kristin J. Dana received the PhD from Columbia University (NY,NY) in 1999 and the MS degree from Massachusetts Institute of Technology in 1992, and a BS degree in 1990 from the Cooper Union (NY,NY).   She is a Full Professor in the Department of Electrical and Computer Engineering at Rutgers University.  Her research interests in computer vision include robot vision, socially cognizant robotics, deep learning, computational photography, illumination modeling, texture and reflectance.  Dr. Dana is also a member of the Rutgers Center for Cognitive Science and a member of Graduate Faculty of the Computer Science Department.   From 1992-1995 she was on the research staff at SRI-Sarnoff Corporation developing real-time motion estimation algorithms. She is the recipient of the  National Science Foundation Career Award (2001) for a program investigating surface science for vision and graphics and a team member recipient of the Charles Pankow Innovation Award in 2014 from the ASCE. Dr. Dana currently leads an NSF National Research Traineeship (NRT) at Rutgers University entitled SOCRATES: Socially Cognizant Robotics for a Technology Enhanced Society. 
\end{IEEEbiography}


%\vfill


\end{document}